\documentclass{IEEEtran}

\usepackage[style=ieee, sorting=none, citestyle=numeric-comp]{biblatex}
\usepackage[%
    colorlinks=True,
    citecolor=blue,
    pdfborder={0 0 0},
    linkcolor=red
]{hyperref}

\usepackage{graphicx}
\graphicspath{ {./Figures/} }

\usepackage{graphics} 
\usepackage{epsfig} 
\usepackage{mathptmx} 
\usepackage{times} 
\usepackage{amsmath} 

\DeclareMathOperator*{\argmax}{arg\,max}
\DeclareMathOperator*{\argmin}{arg\,min}
\usepackage{amssymb}  
\usepackage{bm}
\usepackage{color}

\usepackage{graphicx}

\usepackage{graphics} 

\usepackage{amssymb}
\usepackage{wrapfig}
\usepackage{multirow}
\usepackage{tabularx,booktabs}
\usepackage{algorithm} 
\usepackage{algpseudocode} 

\newcolumntype{C}{>{\centering\arraybackslash}X} 
\usepackage{lipsum} 
\usepackage{mdframed}

\usepackage{amsthm}
\newtheorem{theorem}{Theorem}
\newtheorem{definition}{Definition}    
\newtheorem{remarknn}{Remark}

\newtheorem{assumption}{Assumption}

\begin{document}


\title{Meta Policy Switching for Secure UAV Deconfliction in Adversarial Airspace}


\author{Deepak Kumar Panda and Weisi Guo
\thanks{This work has been submitted to the IEEE for possible publication. Copyright may be transferred without notice, after which this version may no longer be accessible.}
\thanks{This work was supported by the Royal Academy of Engineering and the Office of the Chief Science Adviser for National Security under the UK Intelligence Community Postdoctoral Research Fellowship programme}
\thanks{$^{1}$D. Kumar Panda and W. Guo are with the Faculty of Engineering and Applied Sciences,
Cranfield University, MK43 0AL Cranfield, U.K
        {\tt\small Deepak.Panda@cranfield.ac.uk, weisi.guo@cranfield.ac.uk}}%
}

\markboth{IEEE Transactions on XXXX.}
{Shell \MakeLowercase{\textit{et al.}}: Bare Demo of IEEEtran.cls for IEEE Journals}

\maketitle

\begin{abstract}
Autonomous UAV navigation using reinforcement learning (RL) is vulnerable to adversarial attacks that manipulate sensor inputs, potentially leading to unsafe behavior and mission failure. Although robust RL methods provide partial protection, they often struggle to generalize to unseen or out-of-distribution (OOD) attacks due to their reliance on fixed perturbation settings. To address this limitation, we propose a meta-policy switching framework in which a meta-level policy dynamically selects among multiple robust policies to counter unknown adversarial shifts. At the core of this framework lies a discounted Thompson sampling (DTS) mechanism that formulates policy selection as a multi-armed bandit problem, thereby minimizing value distribution shifts via self-induced adversarial observations. We first construct a diverse ensemble of action-robust policies trained under varying perturbation intensities. The DTS-based meta-policy then adaptively selects among these policies online, optimizing resilience against self-induced, piecewise-stationary attacks. Theoretical analysis shows that the DTS mechanism minimizes expected regret, ensuring adaptive robustness to OOD attacks and exhibiting emergent antifragile behavior under uncertainty. Extensive simulations in complex 3D obstacle environments under both white-box (Projected Gradient Descent) and black-box (GPS spoofing) attacks demonstrate significantly improved navigation efficiency and higher conflict-free trajectory rates compared to standard robust and vanilla RL baselines, highlighting the practical security and dependability benefits of the proposed approach.

\end{abstract}


%
\IEEEpeerreviewmaketitle

\section{Introduction}
With the rising deployment of UAVs in civil airspace, ensuring safe and conflict-free navigation is critical \cite{huang2019collision}, which necessitates robust unmanned aircraft traffic management (UTM) systems \cite{kopardekar2016unmanned}. According to ICAO \cite{organisation2005global}, conflict detection and resolution (CDR) in UTM operates across three layers: \textit{strategic conflict management} (pre-flight deconfliction) \cite{tsourdos23}, \textit{separation provision} (in-flight tactical decisions) \cite{ Huang2023UTM}, and \textit{collision avoidance} (post-conflict mitigation) \cite{yasin2020unmanned}. While strategic deconfliction offers early conflict prevention, it lacks flexibility to handle sudden changes like dynamic obstacles with unknown trajectories. Tactical deconfliction, by contrast, supports real-time adaptation through onboard sensors and surveillance technologies.

Reinforcement learning (RL) has emerged as a promising tool for tactical deconfliction \cite{zhao2021physics, li2024graph}, enabling UAVs to adapt to real-time conditions by learning optimal control policies in high-dimensional state-action spaces. However, RL policies are heavily reliant on sensor data and communication channels, which are vulnerable to adversarial attacks such as spoofing, jamming, and evasion \cite{yu2023cybersecurity, panda2023fragility, Kai24spoof}. These threats not only degrade perception but can also exploit the RL agent’s policy directly \cite{Kai2024RL, wei2024survey}. To address these challenges, this paper explores a \textit{meta-policy} switching framework for UAV tactical deconfliction under adversarial attacks. Rather than merely withstanding disruptions, the meta-policy learns to switch against an ensemble of robust policies against self-induced adversarial observations. Our approach integrates adversarially trained policies with discounted Thompson sampling (DTS) to adaptively switch to the most robust policy in response to unseen perturbations, enhancing both safety and resilience in UTM environments.
\subsection{Related Works}
\subsubsection{Tactical UAV Deconfliction}
Reinforcement learning (RL) offers a scalable alternative for tactical deconfliction over existing planning algorithms by optimizing policies over high-dimensional state spaces to train obtain navigation policy against dynamic obstacles. However, existing RL-based methods often struggle with generalization under adversarial perturbations. To enhance efficiency and safety against 3D obstacles,  techniques such as interfered fluid dynamics systems (IFDS) \cite{yao2015uav, yao2015real}, has been used to incorporate UAV kinematics and reduce path cost \cite{celestini2022trajectory}. Recent work integrating IFDS with deep RL has shown improved resilience and adaptability \cite{zhang2021adaptive}. Building upon this, our approach enhances RL-based deconfliction by incorporating adaptive policy switching to counter adversarial threats.
\subsubsection{Adversarial Threats and Robust RL Defenses}
Adversarial attacks in RL are broadly categorized into perturbations of the action and state space,  policy network parameters, and reward functions \cite{ilahi2021challenges, Kai2024RL, chu2024survey}. State-space attacks use gradient-based loss functions \cite{pattanaik2018robust} or saliency maps \cite{papernot2016limitations} to mislead the trained policies. Action-space and policy network parameter attacks manipulate policy behavior and model integrity \cite{lee2020spatiotemporally, huai2020malicious}, while reward-space attacks distort agent learning \cite{Kai24spoof}. Defensive strategies enhance robustness by preparing the agents against known perturbation bounds which include adversarial training \cite{pattanaik2018robust, behzadan2017whatever}, defensive distillation \cite{papernot2016distillation, delgrange2022distillation}, risk-aware modeling \cite{pan2019risk, tessler2019action}, noisy reward shaping \cite{wang2020reinforcement} and robust adversary-aware RL (RARL) \cite{zhang2021robust, pinto2017robust}. However, the robust RL, trained against bounded perturbations, when exposed to previously unseen  threats, their performance often degrades due to lack of adaptability. 
\subsubsection{Meta-Policy Adaptation}
Meta-policy approaches generally fall into three families: (1) gradient-based meta-RL, (2) policy ensembling with learned switching among low-level policies, and (3) runtime-assurance architectures that switch from advanced to safe controllers. Gradient-based meta-RL algorithms disentangle task inference from control and adapt via latent context inference, improving sample efficiency over on-policy meta-RL, yet they still require online updates and task-posterior estimation at test time, which can be brittle under distribution shift in the state-space \cite{rakelly2019efficient, gao2023context}. While such meta-policies enable fast adaptation across tasks \cite{rakelly2019efficient}, their need for retraining or test-time gradient steps makes them ill-suited for real-time UAV navigation. In offline RL, evaluation-time switching between an RL policy and a behavior-cloning policy using uncertainty estimates has demonstrated adaptation can be done in real-time without retraining \cite{neggatu2025evaluation}, and constraint-adaptive policy switching (CAPS) trains a suite of policies with different reward/cost trade-offs and selects among them at test time to satisfy constraints \cite{chemingui2025constraint}. These advances strengthen the case for evaluation-time switching, but they primarily address dataset coverage and safety-constraint satisfaction rather than adversarial OOD threats, and they do not also provide formal robustness guarantees during switching. Runtime-assurance frameworks such as Black-Box Simplex offer supervisory switching for safety \cite{sheikhi2024black}, yet they do not directly address cyber threats or value-aware adaptation under adversarial manipulation.
\subsection{Research Gap and Contribution} 
\subsubsection{Research Gap}
Prior robust/ensemble RL methods \cite{agarwalcontrastive, osband2016deep} typically average or blend policies to stabilize performance under stochastic variation but do not adapt online to unseen, adversarial shifts. Likewise, adversarially trained approaches \cite{pinto2017robust, mandlekar2017adversarially} presume fixed train-time threat models and lack runtime adaptation. Meta-RL techniques \cite{rakelly2019efficient} enable fast cross-task adaptation but require test-time gradient updates or retraining, which is impractical for real-time UAV operations. Other switching frameworks \cite{neggatu2025evaluation, chemingui2025constraint} demonstrate test-time selection, yet they offer no performance guarantees during switching.
\subsubsection{Contributions}
In this paper, we propose a meta-policy switching framework for UAV deconfliction in adversarial 3D obstacles environment that adaptively switches between pre-trained adversarial action-robust RL policies and a nominal policy.  In contrast to existing research areas of adversarial learning, our method uses discounted Thompson sampling (DTS) driven by value-distribution shifts, quantified via the Wasserstein distance, to select among robust policies without any test-time retraining. Unlike the existing meta-policy adaptive cases, here we provide a theoretical analysis showing how our framework minimizes expected regret while yielding adaptive robustness to attacks which causes OOD shifts in observation space, and we establish a novel connection to antifragility, proving hierarchical policy ensemble leads to emergent antifragile behavior. To our knowledge, this is the first integration of DTS with robust RL ensembles for real-time adaptation in UAV systems against self-induced adversarial observations, accompanied by guarantees for adaptive robustness and antifragility.
\begin{itemize}
	\item \textbf{DTS-Guided Meta-Policy Adaptation:} We introduce a meta-policy switching framework that, to our knowledge, is the first to couple discounted Thompson sampling (DTS) with ensembles of adversarially robust RL policies. The meta-policy uses self-induced adversarial observations and value-distribution (Wasserstein-1) shift estimates to switch policies online, enabling test-time adaptation without retraining.
	\item \textbf{Theoretical Analysis of Adaptive Robustness and Antifragile Behaviour:} We provide a formal regret analysis showing that the DTS sampler asymptotically selects the policy that minimises value-function shifts against self-induced adversarial observations, even under previously unseen perturbations. We further establish conditions under which reduced regret yields \emph{antifragile} improvement i.e., performance increases as adversarial intensity rises, offering, to our knowledge, the first rigorous characterisation of test-time adaptive robustness and antifragility in adversarial RL.
	\item \textbf{Empirical Validation Under Attacks which causes OOD shifts:} Beyond bounded, known attacks, we evaluate against both adversarial attacks which causes low-OOD (projected gradient descent) and high-OOD shift (hijacking-induced GNSS spoofing) in observation space  in deconfliction against 3D obstacles.
	The proposed strategy consistently improves navigation efficiency and conflict-free trajectory rates over robust and vanilla RL baselines, 
	We also performed ablation studies while considering different bandit samplers, and showed that DTS based sampler provided better safety results against the other techniquues.
\end{itemize}
\textbf{Paper Organization:}Section \ref{3DModel} elaborates on UAV deconfliction environment against 3D obstacles. Section \ref{action_robust} provides a comprehensive examination of action-based RL. In Section \ref{AF_DTS}, the formulation of the robust policies as multi-armed bandits are considered, along with the respective Bernoulli rewards. Section \ref{theory} focuses on the optimality of the Bernoulli rewards with respect to the distribution shift considering the regret minimization problem, and the optimality of adaptation during test-time due to DTS sampler while hence inducing antifragile behaviour.  A detailed analysis of the results, including a comparative assessment against robust algorithms based on benchmarks, is provided in Section \ref{results}. Finally, Section \ref{concl} synthesizes the key findings and conclusions.
\section{3D UAV Navigation: Environment and Agent Description} \label{3DModel}
\subsection{State Transition Model}
The modeling environment is a tactical deconfliction challenge, where the UAV has to navigate from the starting point to the goal while avoiding collision against dynamic 3D obstacles. The resulting trajectory is computed by the interaction between the original flow field due to the target and the interfered flow field due to the obstacle(s).  The initial flow field is designed to guide the UAV to the target destination where position vector $\mathbf{P}$, influences the velocity $u\left (  \mathbf{P} \right )$ of the UAV as:
\begin{equation} \label{eq:1}
	u\left (  \mathbf{P} \right ) = -\left [ \frac{C\left ( x - x_d \right )}{\bar{d}\left ( \mathbf{P}, \mathbf{P}_d \right )}, \frac{C\left ( y - y_d \right )}{\bar{d}\left ( \mathbf{P}, \mathbf{P}_d \right )}, \frac{C\left ( z - z_d \right )}{\bar{d}\left ( \mathbf{P}, \mathbf{P}_d \right )}  \right ]^T,
\end{equation}
where $C$ is the attraction constant towards the goal and $\bar{d}$ being the Euclidean distance between current position $\mathbf{P}$ and the target position $\mathbf{P}_d$.
\begin{figure}[thpb]
	\centering
	\includegraphics[scale=0.16]{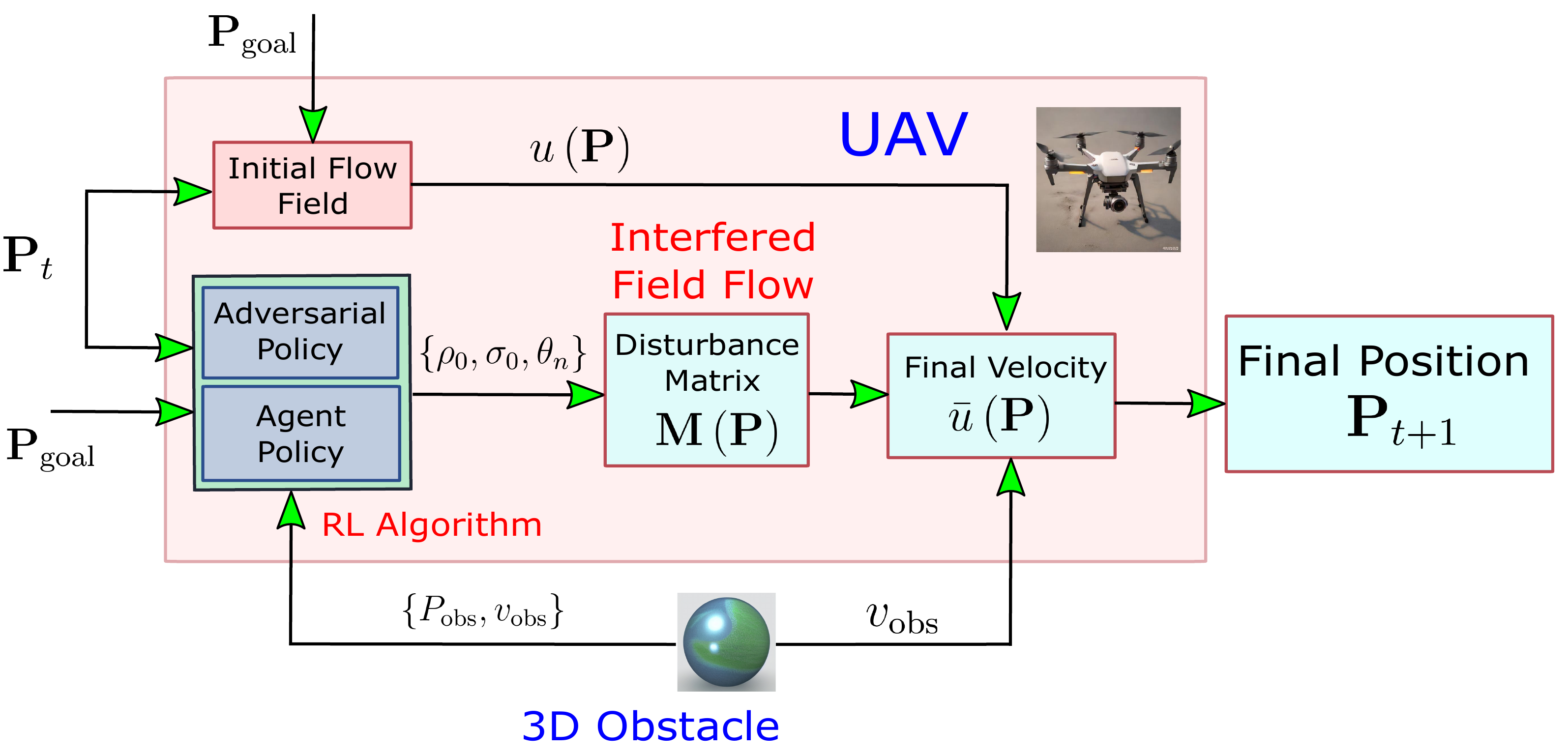}
	\caption{The practical implementation of the RL algorithm for the UAV navigation, whose path is smoothened by IFDS.}
	\label{practical_implementation}
\end{figure}
However, the initial flow field is affected due to the presence of dynamic obstacles. The effect due to the motion and shape of the obstacles is encapsulated using $\overline{M} \left ( \mathbf{P} \right )$ which is derived in Appendix \ref{inter_flow}.  The resulting synthetic disturbance flow field in the dynamic obstacle environment is provided as follows,
\begin{equation} \label{eq:2}
	\overline{u} \left ( \mathbf{P} \right ) = \overline{M} \left ( \mathbf{P} \right ) \left ( u\left ( \mathbf{P} \right ) - v_{\textup{obs}} \left ( \mathbf{P} \right ) \right ) + v_{\textup{obs}} \left ( \mathbf{P} \right ).
\end{equation}
The UAV position $\mathbf{P}_{t+1}$  at time $t+1$ is obtained by using the velocity $\overline{u} \left ( \mathbf{P} \right )$ from (\ref{eq:1}) :
\begin{equation} \label{eq:3}
	\mathbf{P}_{t+1} = \mathbf{P}_{t} + \overline{u} \left ( \mathbf{P}_{t} \right ) \Delta T.
\end{equation}
The manifestation of the transition model on the UAV is shown in Figure \ref{practical_implementation}. The discrete time UAV position update in (\ref{eq:3}) becomes the transition model for the agent which is described in the next subsection.

\subsection{Reinforcement Learning Agent}
The subsequent subsections will explain the state, action, and reward for the RL agent used for developing the navigation policy for a given mission for the UAV.
\subsubsection{States}
The ownship UAV, navigated by an RL agent, acquires data from the closest moving obstacles - potentially another UAV following a predefined trajectory - via ADS-B as part of the UTM infrastructure. This data includes the obstacle's position $\mathbf{P}_{\textup{obs}}$ and its velocity $v^{\textup{obs}}_t$ at time $t$. Consequently, the UAV considers its relative position to both the destination and the obstacle, as well as the obstacle's velocity, as input states. Hence, the agent state $\Phi_t$ at time $t$ is given as: 
\begin{equation} \label{eq:4}
	\Phi_t = \left\{ \mathbf{P}_{\textup{goal}} - \mathbf{P}_{\textup{UAV}} , \mathbf{P}_{\textup{obs}} - \mathbf{P}_{\textup{UAV}}, v^{\textup{obs}}_t  \right\}.
\end{equation}
\subsubsection{Reward}
The first reward mechanism $R_1$ is in the form of penalty signal if the planned navigation point $\mathbf{P}_{t+1}$ is inside the dynamic obstacle, which is represented as:
\begin{equation} \label{eq:5}
	R_1 = \frac{\left\| \mathbf{P}_{t+1} - \mathbf{O}_{\textup{obs}} \right\|}{R_{\textup{obs}}} \; \; \textup{if} \left\| \mathbf{P}_{t+1} - \mathbf{O}_{\textup{obs}}  \right\| \leq R_{\textup{obs}},
\end{equation}
where $\mathbf{O}_{\textup{obs}}$ is considered the center of the obstacle ball, $R_{\textup{obs}}$ is considered its radius. In case the UAV planned path is not inside the dynamic obstacles, the reward mechanism aims to optimize the path length, which can be specified as:
\begin{equation} \label{eq:6}
	R_2 = \left\{\begin{matrix}
		-\frac{\left\| \mathbf{P}_{t+1} - \mathbf{P}_g \right\|}{\left\| P_{\textup{obs}} - \mathbf{P}_g \right\|} \: \: \textup{if} \left\| \mathbf{P}_{t+1} - \mathbf{P}_g \right\| > \varepsilon,  \\
		-\frac{\left\| \mathbf{P}_{t+1} - \mathbf{P}_g \right\|}{\left\| P_{\textup{obs}} - \mathbf{P}_g \right\|} + C_1 \: \: \textup{if} \left\| \mathbf{P}_{t+1} - \mathbf{P}_g \right\| \leq \varepsilon.
	\end{matrix}\right.
\end{equation}
As described in (\ref{eq:6}), the deconfliction is achieved by prioritizing the shortest path to the target. However, the shortest path may bring the UAV closer to the obstacle surface. To address this risk, a threat zone is defined around the obstacle, where a penalty is imposed if the planned path intersects this zone. 
\begin{equation} \label{eq:7}
	\begin{gathered}
		R_3 = \frac{\left\| \mathbf{P}_{t+1} - \mathbf{O}_{\textup{obs}} - \left ( R_{\textup{obs}} + r \right ) \right\|}{R_{\textup{obs}}} - C_2, \; \; \\ \textup{if} \; \;  R_{\textup{obs}} < \left\| \mathbf{P}_{t+1} - \mathbf{O}_{\textup{obs}} \right\| < R_{\textup{obs}} + r.
	\end{gathered}
\end{equation}
As specified in (\ref{eq:7}) $r>0$ is considered a threat area where $C_2$ is a positive constant. Hence, we can scale the above rewards $R_1$, $R_2$ and $R_3$ as,
\begin{equation} \label{eq:8}
	R_t \left ( \left.\begin{matrix}
		a_t \end{matrix}\right| s_t  \right ) = \lambda_1 \cdot  R_1 + \lambda_2 \cdot R_2 + \lambda_3 \cdot  R_3.
\end{equation}
where $\lambda_1, \lambda_2, \lambda_3 > 0$ is considered as a weighting factor.
\subsubsection{Action} The parameters $\left\{ \varrho_0, \varsigma_0, \vartheta_n \right\}$ as defined in  Appendix \ref{inter_flow}  becomes the action space of the UAV agent. 

\begin{remarknn}
	While the simulation environment described in this work incorporates complex 3D obstacles, it does not explicitly model several key real-world factors, such as sensor noise, actuator latency, communication packet loss, and environmental disturbances. These aspects can influence policy execution and overall system performance in real UAV deployments. Nonetheless, simulation-based evaluation remains a necessary and widely adopted first step for benchmarking and developing learning algorithms in autonomous systems. It enables controlled, repeatable, and safe assessment prior to real-world deployment. As part of our future work, we intend to extend our evaluation to include these realistic uncertainties and validate the proposed algorithms in hardware-in-the-loop and physical UAV testbeds, thereby aligning with the practical requirements of secure cyber-physical systems
\end{remarknn}

\section{Action Robust Reinforcement Learning} \label{action_robust}
\begin{figure}[thpb]
	\centering
	\includegraphics[scale=0.30]{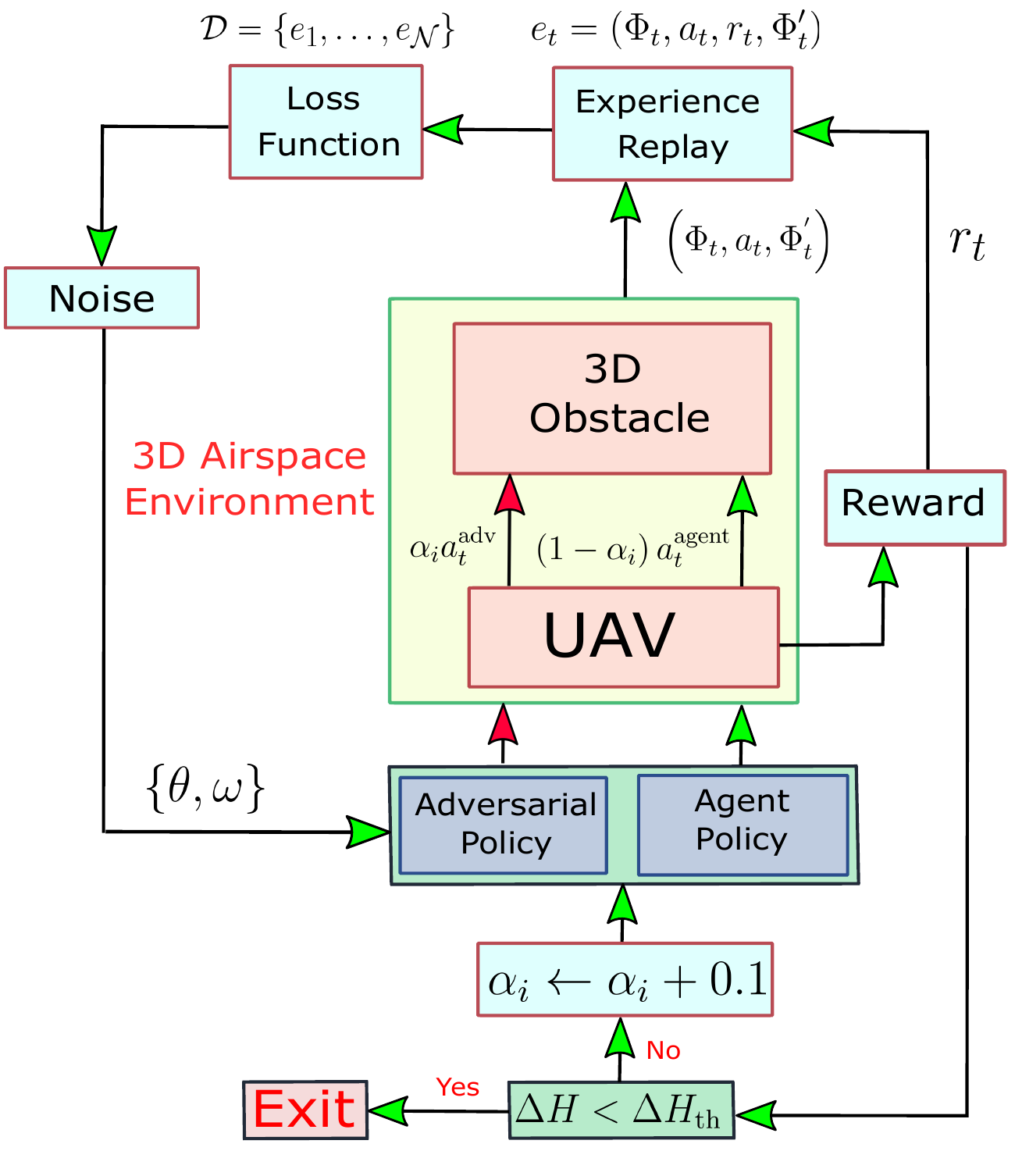}
	\caption{The schematic to train the ensemble adversarial action robust DDPG for various $\alpha_i$ until the convergence is obtained in terms of entropy $\Delta H$.}
	\label{figure2_schematic}
\end{figure}

As illustrated in Figure \ref{figure2_schematic}, it is shown to obtain a set of robust policies by varying the parameter $\alpha$. The agent and adversarial policies, denoted as $\pi_{\theta}$ and $\mu_\omega$, respectively, generate action $a_t$ based on a weighted combination: $\left (1-\alpha_i \right ) a^{\textup{agent}}$ for the agent and $\alpha a^{\textup{adv}}$  for the adversary. These actions determine the trajectory of UAV navigation in 3D obstacles environment. Upon executing an action $a_t$,  the agent receives a reward $r_t$ while transitioning to a new state $\Phi'$. The agent's overall experience is stored as $e_t = \left \{\Phi_t, a_t, r_t, \Phi'_{t} \right  \} $ in the replay buffer $\mathcal{D}_{\textup{replay}}$. Subsequently, a batch of experiences $\mathcal{D} = \left \{e_1, \cdots, e_N \right \}$ is sampled from the replay buffer to compute the critic and the corresponding actor loss function.

To facilitate joint optimization of agent and adversarial policies under a common critic, a min-max objective is formulated. Stochastic gradient descent (SGD) updates are performed with added noise to prevent the Nash equilibrium solution from becoming trapped at saddle points. Once the agent is trained, the difference in entropy between the exploration and exploitation rewards, $\Delta H$, is computed. If $\Delta H$ exceeds a predefined threshold, the training resumes with an updated value $\alpha$. However, if $\Delta H$ falls below the threshold, the training is terminated, resulting in the final set of robust action policies.
In the context of a vanilla RL, we define the Markov decision process (MDP) $\mathcal{M}:= \left ( \mathcal{S}, \mathcal{A}, T_1, \gamma, P_0, R_1 \right )$, where $\mathcal{S}$ and $\mathcal{A}$ represent the state and action spaces, respectively. For addressing RL problems with continuous action spaces, we define it as $\mathcal{A} \in \mathbb{R}^d$. The transition dynamics is given by $T_1 : \mathcal{S} \times \mathcal{S} \times \mathcal{A} \rightarrow \left [ 0,1 \right ]$ where $T_1  \left ( \left.\begin{matrix}
	\Phi_t^{'} \end{matrix}\right| \Phi_t, a_t \right )$ denotes the probability of reaching the state $\Phi^{'}_t$ by taking the action $a_t$ at the state $\Phi_t$. $P_0 : \mathcal{S} \rightarrow  \left [ 0, 1 \right ]$ represent the initial distribution on $\mathcal{S}$. $\gamma$ is considered as the discount factor $R_1 : \mathcal{S} \times \mathcal{A} \rightarrow \mathbb{R}$ represents the reward of the environment after taking the action $a_t$.

However, to train agent and adversarial policies in a shared action space, a two-player zero-sum Markov game is used \cite{tessler2019action}.  Zero-sum Markov game provides a principled framework for training robust agents by explicitly modeling an adversary that aims to minimize the agent’s reward. This formulation captures worst-case interactions between the agent and its environment, enabling the learning of policies that are resilient to strategically optimized disruptions. Unlike methods based on fixed perturbation bounds or stochastic noise, adversarial training in a zero-sum setting simulates dynamic threats based on worst-case vulnerabilities of the agent, thereby anticipating and minimizing potential reward degradation. 

In this setting, both players simultaneously select an action and the reward is determined by the current state $\Phi_t$ and the joint actions of the agent and adversarial policies, which influences the transition dynamics. The MDP models the game as $\mathcal{M}_2 = \left ( \mathcal{S}, \mathcal{A}, \mathcal{A}', T_2, \gamma, R_2, P_0 \right ) $ where $\mathcal{A}$ and $\mathcal{A}'$ denote the action spaces available to the agent and the adversary. The state transition probability is given by $T_2: \mathcal{S} \times \mathcal{A} \times \mathcal{A}' \times \mathcal{S} \rightarrow  \mathbb{R}$ and $R_2: \mathcal{S} \times \mathcal{A} \times \mathcal{A}' \rightarrow \mathbb{R}$ defines the reward for both players. Here, the agent's policy is represented by $\mu: \mathcal{S} \rightarrow \mathcal{A}$, while the adversary follows the policy $\nu: \mathcal{S} \times \mathcal{A}'$ within the environment $\mathcal{M}_2$. At each time step $t$, the agent and the adversary select actions $a_t = \mu \left ( \Phi_t \right )$ and ${a_t}' = \nu \left ( \Phi_t \right )$, respectively. In the context of a zero-sum game, the agent receives a reward $r_t = R_2 \left ( \Phi_t, a_t, a^{'}_t \right )$, while the adversary receives a reward of $-r_t$. In the adversarial game, the following performance objective is considered for optimization. 
\begin{equation} \label{eq:9}
	J\left ( \mu, \nu \right ) = \mathbb{E} \left [  \left.  \begin{matrix} \sum_{t=1}^{\infty} \gamma^{t-1} r_t 
	\end{matrix}\right| \mu, \nu, \mathcal{M}_2 \right ],
\end{equation}
where $\sum_{t=1}^{\infty} \gamma^{t-1} r_t$ is the random cumulative return. If we consider parameterized policies $\left\{ \mu_{\theta}: \theta \in  \Theta  \right\}$ and $\left\{ \nu_{\omega}: \omega \in  \Omega  \right\}$ for the agent and adversary policy respectively, the reward maximization objective is given as:
\begin{equation}\label{eq:10}
	\max_{\theta \in \Theta} \min_{\omega \in \Omega} J\left ( \mu_{\theta}, \nu_{\omega} \right ). 
\end{equation}
The objective $J$ in (\ref{eq:10}) is non-convex and non-concave with respect to both $\theta$ and $\omega$. To address this, the analysis considers the set of all probability distributions over $\Theta$ and, $\Omega$ aiming to identify the optimal distribution that solves the following objective: 
\begin{equation}\label{eq:11}
	\max_{p \in \mathcal{P} \left ( \Theta \right )} \min_{q \in \mathcal{P} \left ( \Omega \right )}  f\left ( p,q \right ) := \mathbb{E}_{\theta \sim p} \left [ \mathbb{E}_{\omega \sim q} \left [ J \left ( \mu_{\theta}, \nu_{\omega} \right ) \right ]  \right ].
\end{equation}
The objective defined in (\ref{eq:11}) can be solved using stochastic gradient Langevin dynamics as shown in Appendix \ref{langevin}.
To determine the number of models in the ensemble set of agent and adversarial policies, entropy is employed as a metric. According to \cite{han2021max, eysenbach2022maximum}, the goal to achieve robustness in RL is to minimize the maximum entropy of environmental states. With the increase in adversarial policy's contribution to the action space, we expect the entropy difference between the exploitation and exploration phases of learning to decrease. However, at higher levels of adversarial contribution, states may exhibit increased entropy, leading to greater randomness in episodic rewards. Let $\mathbf{H}_{\textup{rand}}$ represent the reward entropy during the exploration phase and $\mathbf{H}_{\textup{opt}}$ during the exploitation phase; then, the difference in entropy can be expressed as $\Delta \mathbf{H}  = \mathbb{E} \left [ \Delta \mathbf{H}^{\textup{rand}}_{\alpha_i} - \max \left [ \mathbf{H}^{\textup{opt}}_{\alpha_i}  \right ]   \right ]$. As shown in Figure \ref{figure2_schematic}, $\Delta \mathbf{H}$ is used as a metric to find the required number of ensembles, where, if it goes below, $\Delta H_{\textup{th}}$ then we do not consider the given model within the ensemble set.
To determine the number of models in the ensemble set of agent and adversarial policies, entropy is employed as a metric. According to \cite{han2021max, eysenbach2022maximum}, the goal to achieve robustness in RL is to minimize the maximum entropy of environmental states. With the increase in adversarial policy's contribution to the action space, we expect the entropy difference between the exploitation and exploration phases of learning to decrease. However, at higher levels of adversarial contribution, states may exhibit increased entropy, leading to greater randomness in episodic rewards. Let $\mathbf{H}_{\textup{rand}}$ represent the reward entropy during the exploration phase and $\mathbf{H}_{\textup{opt}}$ during the exploitation phase; then, the difference in entropy can be expressed as $\Delta \mathbf{H}  = \mathbb{E} \left [ \Delta \mathbf{H}^{\textup{rand}}_{\alpha_i} - \max \left [ \mathbf{H}^{\textup{opt}}_{\alpha_i}  \right ]   \right ]$. As shown in Figure \ref{figure2_schematic}, $\Delta \mathbf{H}$ is used as a metric to find the required number of ensembles, where, if it goes below, $\Delta H_{\textup{th}}$ then we do not consider the given model within the ensemble set. The entire algorithm to obtain the ensemble of robust policies is shown in Algorithm \ref{Robust_DDPG}.

\section{Discounted Thompson Sampling for Switching Robust Policies} \label{AF_DTS}
\begin{figure}[thpb]
	\centering
	\includegraphics[scale=0.22]{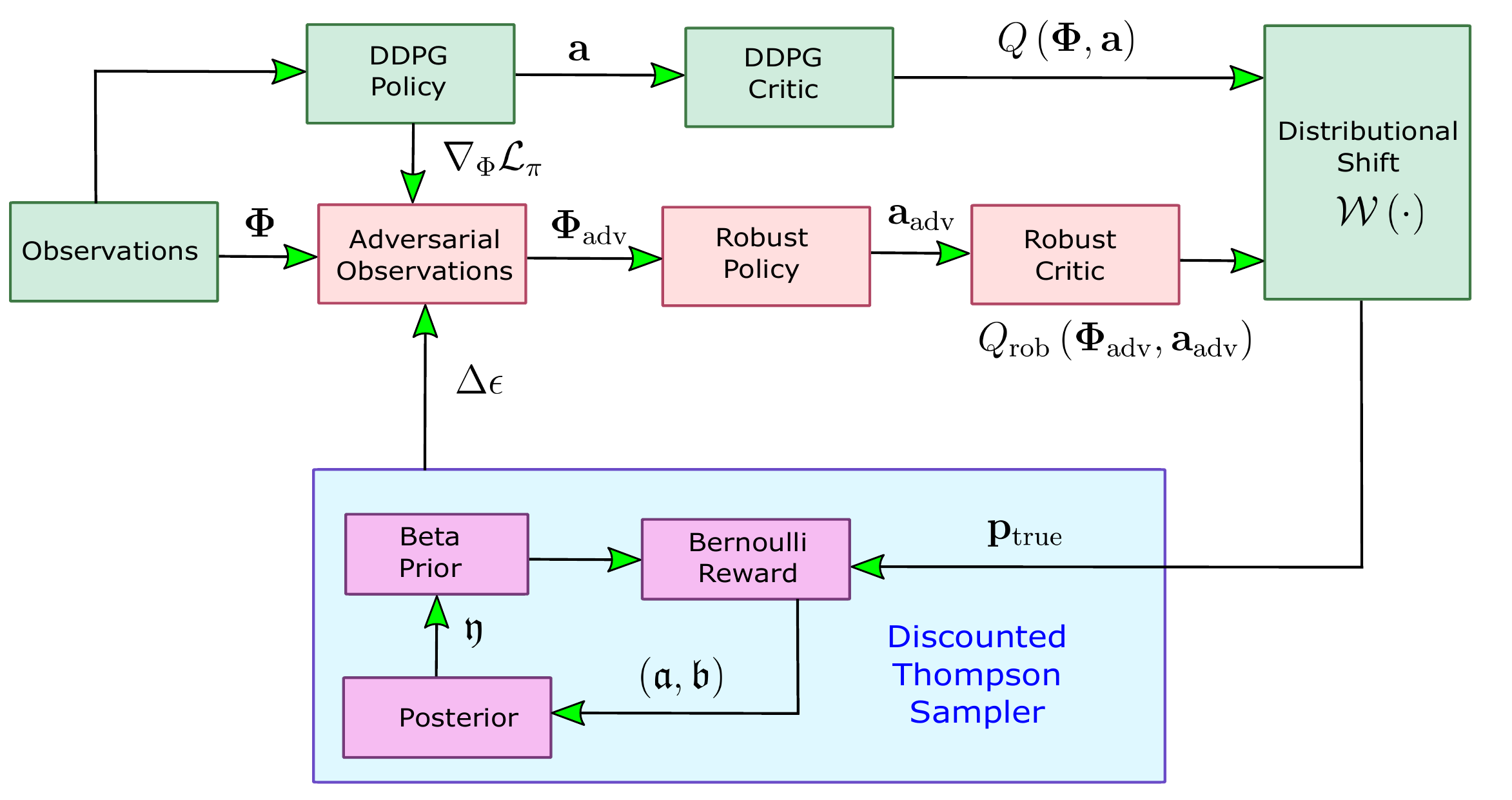}
	\caption{Meta-policy design using Discounted Thompson sampling (DTS) to select the robust policies against self-induced adversarial observations.}
	\label{figure4_schematic}
\end{figure}
As illustrated in Figure \ref{figure4_schematic}, we present the novel DTS-based adaptive policy-switching framework for adaptive robustness against unseen adversarial attacks. This method uniquely addresses the limitation of the existing adaptive adversarial RL algorithms \cite{pinto2017robust, rakelly2019efficient}  ,which are computationally expensive and impractical in real-time UAV operations, as it dynamically selects among pre-trained robust policies using discounted DTS based on real-time estimation of the value distribution shifts thus offering efficient, real-time adaptive robustness against adversarial perturbations without any retraining during deployment.

Adversarial observations, denoted as $\Phi_{\textup{adv}}$, are synthesized from the original observations $\Phi$ by incorporating the gradient sign of DDPG actor loss, $\nabla_{\Phi} \mathcal{L}_\pi$, with the maximum magnitude of the perturbation $\epsilon$. Using the robust action policy, as described in the previous section, the corresponding adversarial action $\mathbf{a}_{\textup{adv}}$ is obtained. Similarly, from the nominal policy $\pi$ and original observations $\Phi$, the agent actions $\mathbf{a}$ are obtained. To assess the impact of adversarial observations, the value distributions are obtained from trained critics, specifically $Q(\Phi, \mathbf{a})$ for vanilla DDPG and $Q_{\textup{rob}}(\Phi_{\textup{adv}}, \mathbf{a}_{\textup{adv}})$ for action robust DDPG. The distributional shift is quantified using the Wasserstein distance between these value functions. Subsequently, the distributional shift is normalized with respect to the minimum distance for all robust value functions to compute the true probability ${p}_{\textup{true}}$.

With the selected robust policy, the Bernoulli reward, which represents the likelihood, is calculated based on a Beta prior with parameters $(\mathfrak{a}, \mathfrak{b})$, while updating the conjugate Beta posterior distribution. After a certain number of time steps, once the posterior update for the DTS is complete, the self-induced attack strength is increased by $\Delta \epsilon$, resulting in a new set of probabilities, $\mathbf{p}_{\textup{true}}$, thus addressing the nonstationary Bernoulli reward. Consequently, the DTS parameters $(\mathfrak{a}, \mathfrak{b})$ are updated with a discount factor $\mathfrak{y}$, which accounts for the temporal variations in $\mathbf{p}_{\textup{true}}$. The methodology for generating self-induced adversarial observations using the Frank-Wolfe optimization technique is presented in the following subsection.
\subsection{Self-Induced Adversarial Observations}
Here, the meta-policy switching adaptation is obtained using self-induced adversarial observations to ensure robustness. The adversarial observations are called self-induced in this context, as it is generated from nominal policy $\pi$ known to the agent. Here, a projection-free self-induced adversarial attack mechanism based on the Frank-Wolfe algorithm \cite{chen2020frank} is used. This approach leverages an iterative first-order white-box attack with momentum, which exhibits a convergence rate of $O \left ( 1 \mathbin{/} \sqrt{T} \right )$, where $T$ represents the number of steps. For a given state representation $\Phi$, let $\mathcal{P}_{\mathcal{X}} \left ({\Phi} \right)$ denote the projection operation that maps the observation $\Phi$ to the constraint set $\mathcal{X}$ at each iteration. In contrast to the projected gradient descent attack (PGD) \cite{gupta2018cnn}, the Frank-Wolfe algorithm is inherently projection-free, as it instead relies on a linear minimization oracle (LMO) over the constraint set $\mathcal{X}$ at each iteration. The LMO is defined as follows, 
\begin{equation} \label{eq:12}
	\textup{LMO} \in \argmin_{\Phi_{\textup{adv}} \in \mathcal{X}}  \langle {\Phi}_{\textup{adv}}, \nabla \mathcal{L}_{\pi} \left ({\Phi} \right) \rangle.
\end{equation}
The LMO in (\ref{eq:12}) is seen as the minimization of the first-order Taylor expansion of $\mathcal{L}_{\pi}$ at $\Phi_t$:
\begin{equation}\label{eq:13}
	\min_{{\Phi}_{\textup{adv}} \in \mathcal{X}} \mathcal{L}_{\pi} \left ( {\Phi} \right ) + \langle {\Phi}_{\textup{adv}} - {\Phi}, \nabla \mathcal{L}_{\pi} \left ({\Phi} \right) \rangle.
\end{equation}
The minimization procedure to obtain an self-induced adversarial state in (\ref{eq:13}) can be implemented using the process shown in Algorithm \ref{alg:adv_state}.
In the Frank-Wolfe algorithm, the linear minimization oracle (LMO) optimally solves a linear subproblem in the constraint set,$\mathcal{X}$, followed by a weighted averaging step with the previous iterate to compute the final update. Compared to the PGD attack, the Frank-Wolfe approach is more conservative, as it eliminates the need for explicit projection steps, resulting in improved self-induced distortion characteristics.
\subsection{Distributional Shift Due to Adversarial Observation}
The shift in the value distribution caused by self-induced adversarial observations is quantified using the 1-Wasserstein norm, which is normally used to measure the distance between probability distributions. Let $\mathcal{P} \left (\mathbb{R}^d \right)$ denote the space of probability measures in $\mathbb{R}^d$ with finite first moments. Given two probability measures, $\beth, \daleth \in \mathcal{P} \left (\mathbb{R}^d \right)$, the 1-Wasserstein is defined as:
\begin{equation} \label{eq:14}
	\mathcal{W}_1 \left (\beth, \daleth \right ) = \inf_{\iota \in \Pi \left( \mu, \nu \right )} \mathbb{E}_{\left ( x,y \sim \iota \right )} \left [ \left\| x - y\right\| \right ].
\end{equation}
Here $\Pi \left ( \beth, \daleth \right )$ represents the set of all joint distributions $\iota$ with marginals $\beth$ and $\daleth$. $\left\| x - y\right\|$ represents the Euclidean norm in $\mathbb{R}^d$, and $\mathbb{E}_{\left ( x,y \sim \iota \right )} \left [ \left\| x - y\right\| \right ]$ quantifies the cost of transporting mass from $x$ to $y$. Similarly, if we consider a critic for vanilla DDPG be $Q$ and for action robust critic as $Q^{\alpha}_{\textup{AR}}$, for which we can define the value distribution $Z \left ( \Phi \right )$ and $Z^{\alpha}_{\textup{AR}} \left ( \Phi_{\textup{adv}} \right )$ follows:
\begin{equation} \label{eq:15}
	\begin{gathered}
		Z \left ( \Phi \right ) = Q_{\textup{DDPG}} \left ( \Phi, \pi \left ( \Phi \right ) \right ), \\
		Z^{\alpha}_{\textup{AR}} \left ( \Phi \right ) = Q^{\alpha}_{\textup{AR}} \left ( \Phi_{\textup{adv}}, \alpha \cdot \pi_{\textup{adv}} \left ( \Phi \right )  + \left ( 1 - \alpha\right ) \cdot \pi_{\textup{agent}}\right ). \\
	\end{gathered}
\end{equation}
We define the distribution shift $d^{\alpha}_{\epsilon}$ due to adversarial observation $\Phi_{\textup{adv}}$ with the intensity $\epsilon$ as,
\begin{equation} \label{eq:16}
	d^{\alpha}_{\epsilon} = \inf_{\gamma \in \Pi \left (Z, Z^{\alpha}_{\textup{AR}} \right )} \mathbb{E}_{\left (x,y \right ) \sim \gamma} \left [  \left | x-y \right | \right].
\end{equation}
Similarly for DDPG policy,$\pi_{\textup{DDPG}}$, we define the value distribution at adversarial observation $\Phi_{\textup{adv}}$ as,
\begin{equation} \label{eq:17}
	Z_{\textup{adv}} = Q \left ( \Phi_{\textup{adv}}, \pi_{\textup{DDPG}} \left ( \Phi_{\textup{adv}}\right ) \right ).
\end{equation}
The value distribution shift for the DDPG due to adversarial observation $\Phi_{\textup{adv}}$ of magnitude, $\epsilon$ i.e. $d_{\epsilon}$ is defined as,
\begin{equation} \label{eq:18}
	d_{\epsilon} = \inf_{\iota \in \Pi \left (Z, Z_{\textup{adv}} \right )} \mathbb{E}_{\left (x,y \right ) \sim \iota} \left [  \left | x-y \right | \right].
\end{equation}
Thus, for an adversarial observation with a given perturbation magnitude $\epsilon$, the resulting distribution shift can be analyzed for both the standard DDPG and action-robust policies under varying levels of policy perturbation $\alpha$ as follows:
\begin{equation} \label{eq:19}
	\mathbf{d} = \left \{d^{0}_{\epsilon},d^{\alpha_1}_{\epsilon}, \cdots,  d^{\alpha_n}_{\epsilon}\right \}.
\end{equation}
However, to incorporate Bernoulli rewards for DTS, the distribution shifts must be transformed into equivalent Bernoulli-distributed rewards.  Let $d_{\textup{min}} = \min \left ( \mathbf{d} \right )$, denote the minimum distribution shift. Using this, we can define the corresponding parameter for the Bernoulli rewards as follows:
\begin{equation} \label{eq:20}
	\mathbf{p}_{\textup{true}} = \left \{ \frac{k_{\textup{mult}} \cdot {d_{\textup{min}}}  }{d^{0}_{\epsilon}}, \frac{k_{\textup{mult}} \cdot {d_{\textup{min}}}  }{d^{\alpha_1}_{\epsilon}}, \cdots, \frac{k_{\textup{mult}} \cdot {d_{\textup{min}}}  }{d^{\alpha_n}_{\epsilon}}   \right \}.
\end{equation}
According to (\ref{eq:20}), a lower value of $d^{\alpha}_{\epsilon}$ for a given robust model indicates a smaller shift in the state-action value distribution under a self-induced adversarial observation. Consequently, a higher value $p_{\textup{true}}$ for a specific model enhances its likelihood of successful deployment in adversarial settings. Given that $d_{\text{min}} < d^{\alpha}_{\epsilon}$ ,  we can expect that $\mathbf{p}_{\text{true}}$ is bounded within $\left (0, 1 \right )$ if we consider $k_{\text{mult}} \in \left(0, 1 \right )$. The inverse relationship between $d^{\alpha}_{\epsilon}$ and $\mathbf{p}_{\text{true}}$ enables the DTS to prefer policies that induce minimal value distribution shift under attack, thus enhancing adaptive resilience. 


\subsection{Discounted Thompson Sampling for Minimum Distribution Shift}
Let $\mathcal{K} = \left \{ \pi_0, \pi_1, \cdots, \pi_n \right \}$ represent the ensemble set of robust policies as multi-armed bandits (MAB) available to counter an adversarial observation. Given a finite time horizon $T$, the DTS must select a policy network $\pi_k \in \mathcal{K}$ at each time step $t \in \left \{1, \cdots, T  \right \}$.  Let $X_{k,t}$ denote the reward obtained by employing the $k$-th robust policy at time $t$. Consequently, the instantaneous reward $X_{k,t}$ follows a Bernoulli distribution with an expected value of $\grave{\mu}_{k,t} = \mathbb{E} \left [ X_{k,t}\right] $. Therefore, the optimal policy at any given time $t$  is the one that maximizes the expected reward, expressed as $\mathbf{\grave{\mu}}^*_{t} = \max_{k \in \mathcal{K}} \left \{ \grave{\mu}_{k,t} \right \}$.

In non-stationary environments, the expected Bernoulli rewards evolve over time, either following a gradual trend or undergoing abrupt changes. The sequence of rewards for a specific policy $\pi_k$ is denoted by $\grave{\mu}^k = \left \{\grave{\mu}_{k,t}  \right \}_{t=1}^{T}$, while the general sequence of rewards across all policies is represented as $\grave{\mu} = \left \{ \grave{\mu}^k \right \}_{k \in \mathcal{K}}$. If we consider $\mathcal{P}$ to be the family of admissible policies for DTS, let $\varkappa \in \mathcal{P}$ represent a candidate policy responsible for selecting an appropriate robust policy in response to adversarial observations. At each time step $t$, the policy $\varkappa$ chooses the robust policy $I^{\varkappa}_t$ based on an initial prior $U$ and past switching decisions $\left \{ X_{I^{\varkappa}_t, n} \right \}_{n=1}^{t-1}$. Hence, the DTS switching policy $\varkappa_t$ at time $t$ is given by,
\begin{equation}\label{eq:21}
	\varkappa_t = 
	\begin{cases} 
		\varkappa_1(U) & ; \, t = 1 \\ 
		\varkappa_t(U, X_{I_{\varkappa_1,1}}, \ldots, X_{I_{\varkappa_{t-1}, t-1}}) & ; \, t \geq 2. 
	\end{cases}
\end{equation}
Thompson Sampling (TS) \cite{agrawal2012analysis} operates by maintaining a prior distribution over the success probabilities of each Bernoulli-based robust policy and sampling from this prior to determine which robust model to deploy. The prior distribution is typically modeled as a Beta distribution $\mathfrak{B} \left (\mathfrak{a}, \mathfrak{b} \right )$, with the parameters $ \left \{\mathfrak{a}, \mathfrak{b} \right \}$ updated based on the Bernoulli rewards obtained from the deployment of the robust model against adversarial observations. 

In DTS \cite{wu2018learning}, the central concept is to adjust the algorithm to increase the variance of the prior distribution for unexplored robust models, thus improving the probability of deploying these models in response to adversarial observations.
Consider $\mathfrak{a}_0$ and $\mathfrak{b}_0$, denote the initial values of the parameters for the Beta prior distributions, and $\mathfrak{S}_k$ and $\mathfrak{F}_k$ represent the parameters of the posterior Beta distribution, which are discounted before being updated. Consequently, the update equation for $\left \{ \mathfrak{S}, \mathfrak{F} \right \}$ can be expressed as:
\begin{equation}
	\label{eq:22}
	\begin{gathered}
		\mathbb{E} \left [\mathfrak{S}_{k, t+1} \right ] = \mathfrak{y} \mathbb{E} \left [ \mathfrak{S}_{k, t} \right ] + \breve{\mu}_{k,t} \mathbb{P} \left ( I^{\varkappa}_t = k \right ), \\
		\mathbb{E} \left [ \mathfrak{F}_{k, t+1} \right ] = \mathfrak{y}  \mathbb{E} \left [ \mathfrak{F}_{k, t} \right ] + \left (1- \breve{\mu}_{k,t} \right ) \mathbb{P} \left ( I^{\varkappa}_t = k \right ).
	\end{gathered}
\end{equation}
where, $\mathbb{P} \left ( I^{\varkappa}_t = k \right )$  is the probability of selecting the robust model $k$ at instant $t$ with the existing Bernoulli reward history $\left \{ \mathfrak{r}_{I^{\varkappa}_t,n} \right \}_{n=1}^{t-1}$. If we neglect the effect of $\mathfrak{a}_0$ and $\mathfrak{b}_0$, for the robust model not deployed at the instant, we consider the posterior mean as, 
\begin{equation} \label{eq:23}
	\breve{\mu}_{k,t+1} = \frac{\mathfrak{S}_{k,t+1}}{\mathfrak{S}_{k,t+1} + \mathfrak{F}_{k,t+1}} = \frac{\mathfrak{y} \cdot \mathfrak{S}_{k,t}}{\mathfrak{y} \cdot \mathfrak{S}_{k,t} + \mathfrak{y} \cdot \mathfrak{F}_{k,t}} = \breve{\mu}_{k,t}.
\end{equation}
We can write the variance using (\ref{eq:23}) as, 
\begin{equation} \label{eq:24}
\begin{gathered} 
	\breve{\sigma}^2_{k,t+1} = \frac{\mathfrak{S}_{k,t+1} \cdot \mathfrak{F}_{k,t+1}}{(\mathfrak{S}_{k,t+1} + \mathfrak{F}_{k,t+1})^2 (\mathfrak{S}_{k,t+1} + \mathfrak{F}_{k,t+1} + 1)}, \\
	= \frac{\breve{\mu}_{k,t} (1 - \breve{\mu}_{k,t})}{\mathfrak{y} \cdot \mathfrak{S}_{k,t} + \mathfrak{y} \cdot \mathfrak{F}_{k,t} + 1} 
	\geq \frac{\breve{\mu_{k,t}} (1 - \breve{\mu_{k,t}})}{\mathfrak{S}_{k,t} + \mathfrak{F}_{k,t} + 1}.
\end{gathered}
\end{equation}
We observe in (\ref{eq:24}) that the discounting increases the variance of the prior distribution, which in-turn help tackle the non-stationarity of the rewards. 
\subsection{Test-Time Adversarial Attacks} \label{test:attacks}
To evaluate the adaptive robustness of the DTS-based policy switching mechanism, we test it against two distinct and practically motivated classes of adversarial threats: (1) \textit{observation-space adversarial attacks} induced via gradient-based perturbations, and (2) \textit{signal-level attacks} via GNSS spoofing, which introduce physical misalignments in the UAV’s perceived position. This dual-threat framework captures both control-layer and sensor-layer vulnerabilities relevant to dependable cyber-physical systems. The practicality of  test-time threats is elaborated in Appendix \ref{test-time_threat}.
\section{Theoretical Analysis} \label{theory}
In this section we will theoretically prove that DTS based robust policy switching can ensure adaptive robustness by minimizing the regret and also ensure antifragile behaviour. 
\subsection{Regret Minimization}
Before we consider the regret minimization analysis while deploying the optimal robust policy, we consider the following assumption. 
\begin{assumption} 
	The expected return of any policy $\pi_k$ is is Lipschitz-continuous in the Wasserstein distance between clean and adversarial value distributions:
	\begin{equation}\label{eq:25}
		\left| \mathbb{E}[R(\pi_k; \Phi_t^{\text{adv}})] - \mathbb{E}[R(\pi_0; \Phi_t)] \right| \leq L \cdot d_k^{(t)},
	\end{equation}
	where $L > 0$ is a constant
\end{assumption}
The assumption stated in (\ref{eq:25}) takes into account the engineering realism of non-sudden shift in the reward due to self-induced adversarial states. Let us define a proxy reward to assess robustness, similar to (\ref{eq:20}), which assigns higher reward to policies which exhibits smaller value shifts under perturbation.
\begin{equation} \label{eq:26}
	r_k^{(t)} := \frac{d_{\min}^{(t)}}{d_k^{(t)}} \in (0, 1], \quad \text{with} \quad d_{\min}^{(t)} := \min_j d_j^{(t)}.
\end{equation}
Hence if we consider $\pi_t^* := \arg\min_k d_k^{(t)}$ as the optimal policy for DTS with minimal shift under perturbation, the instantaneous regret is given by:
\begin{equation} \label{eq:27}
	\mu_t^* - \mu_{k_t}^{(t)} \leq L \cdot \left( d_{k_t}^{(t)} - d_{\min}^{(t)} \right) = L \cdot \left( \frac{1}{r_{k_t}^{(t)}} - 1 \right) d_{\min}^{(t)}.
\end{equation}
As we sum over a time horizon $T$, hence we obtain the overall regret $\mathcal{R} \left (T \right ) $,
\begin{equation} \label{eq:28}
	\mathcal{R}(T) := \sum_{t=1}^{T} (\mu_t^* - \mu_{k_t}^{(t)}) \leq L \sum_{t=1}^{T} \left( \frac{1}{r_{k_t}^{(t)}} - 1 \right) d_{\min}^{(t)}.
\end{equation}
The relation in (\ref{eq:28}) shows that minimizing the regret is equivalent to selecting the DTS policy with the minimal value shift which justifies the use of $d^{\left (t \right )}_{k}$ in (\ref{eq:19}) as a robustness metric.
As shown in \cite{wu2018learning}, DTS achieves sublinear regret in multi-armed bandit problems with $M$ changepoints, shown as 
\begin{equation} \label{eq:29}
	\mathbb{E}[R(T)] \leq O\left(\sqrt{KMT \log T}\right).
\end{equation}
This analysis is analogous as the changepoints represent the time-instants where the adversarial attack strength was increased. Substituting into our bound in (\ref{eq:29}), we get,
\begin{equation}\label{eq:30}
	\mathbb{E} \left[ \sum_{t=1}^{T} \left( \frac{1}{r_{k_t}^{(t)}} - 1 \right) d_{\min}^{(t)} \right] \leq \frac{1}{L} \cdot O \left( \sqrt{KMT \log T} \right),
\end{equation}
where $\kappa_t$ is the policy selected by DTS at time $t$. As the regret is to be minimized with respect to non-stationary self.induced adversarial observations, now we highlight the adaptive robustness and antifragile behaviour of the meta-policy switching framework. 
\subsection{Adaptive Robustness against Unseen Adversarial Attacks} 
\begin{theorem}[Adaptive Robustness of DTS ] \label{th:1}
	Let $\mathcal{K} = \{ \pi_1, \dots, \pi_K \}$ be a finite set of robust policies, each associated with a reward proxy $r_k^{(t)} := \frac{d_{\min}^{(t)}}{d_k^{(t)}} \in (0,1]$, where $d_k^{(t)}$ is the Wasserstein-1 distance between the clean and adversarial value distributions under policy $\pi_k$ at time $t$. We assume the Lipschitz nature of the reward function, and consider a unique optimal policy under each stationary segment. Then, the DTS policy selector $\kappa_t$ converges in probability to selecting the optimal policy $k^*$ within each stationary segment. Then we can consider,
    \begin{equation} \label{eq:31}
	\lim_{t \to \infty} \mathbb{P}(\kappa_t = k^*) = 1.
	\end{equation}
\end{theorem}
\begin{IEEEproof}
	Consider a stationary interval $I = [\tau_s, \tau_{s+1})$ during which the adversarial perturbation remains fixed, and the reward proxies $r_k^{(t)}$ are i.i.d. samples from a Bernoulli distribution with mean $\mu_k$ for each policy $\pi_k$. as we know the DTS maintains Beta posteriors over $r_k^{(t)}$:
	\begin{equation}\label{eq:32}
	\tilde{r}_k^{(t)} \sim \text{Beta}(S_k^{(t)} + a_0, F_k^{(t)} + b_0).
	\end{equation}
	Here $a_0, b_0$ are the hyperparameter for the prior. Because the updates use discounted sufficient statistics, we can consider as per \cite{wu2018learning},
	\begin{equation} \label{eq:33}
	S_k^{(t)} = \sum_{i=1}^{t} \gamma^{t-i} r_k^{(i)}, \quad F_k^{(t)} = \sum_{i=1}^{t} \gamma^{t-i} (1 - r_k^{(i)}).
	\end{equation}
	which gives a geometrically weighted moving average of the observed rewards. From the standard concentration inequalities \cite{wu2018learning}, the posterior mean $\mathbb{E}[\tilde{r}_k^{(t)}]$ converges to the true reward mean $\mu_k$ as $t \to \infty$, with posterior variance decaying as $O((1 - \gamma)^2)$. 
	Let us consider $k^* = \arg\max_k \mu_k$. The probability that DTS selects a suboptimal policy $j \ne k^*$ is:
	\begin{equation} \label{eq:34}
	\mathbb{P}(\tilde{r}_j^{(t)} > \tilde{r}_{k^*}^{(t)}).
	\end{equation}
	Using concentration bounds for Beta-Binomial tail bounds \cite{agrawal2017near}, this probability decays exponentially:
	\begin{equation} \label{eq:35}
	\mathbb{P}(\tilde{r}_j^{(t)} > \tilde{r}_{k^*}^{(t)}) \leq \exp\left( -\frac{(\mu_{k^*} - \mu_j)^2}{C(1 - \gamma)^2} \right),
	\end{equation}
	for a constant $C$ depending on the reward range and prior.
	For each of the $M$ change points in the adversarial distribution, a new stationary segment with a different optimal policy may emerge. Provided each segment is sufficiently long, DTS will adapt its sampling distribution by concentrating its posterior on the new $k^*$ in each segment.
	Let $\Delta_s = \min_{j \neq k_s^*} (\mu_{k_s^*} - \mu_j)$ be the minimum reward gap in the segment where it is considered stationary, the regret is bounded by $O \left( \frac{K}{\Delta_*^2} \log T \right)$ and the total regret over all segment satisfies:
	\begin{equation} \label{eq:36}
		\mathbb{E}[R(T)] \leq O\left(\sqrt{KMT \log T}\right).
	\end{equation}
	Hence we can conclude that the policy with the minimal value distribution shift yields the highest proxy reward $r^{(t)}_k$ and the DTS posterior concentrates on the arms with the highest mean, which follows the following results.
	\begin{equation} \label{eq:37}
	\boxed{ \lim_{t \to \infty} \mathbb{P}(\kappa_t = \arg\min_k d_k^{(t)}) = 1 }
	\end{equation}
\end{IEEEproof}
\begin{remarknn}[Adaptation to Unseen Adversarial Attacks]
	Theorem \ref{th:1} implies that the DTS mechanism enables robust policy adaptation even in the presence of previously unseen or out-of-distribution (OOD) adversarial perturbations. Although the policy ensemble $\mathcal{P} = \{ \pi_{\alpha_1}, \dots, \pi_{\alpha_K} \}$ is trained only on a discrete set of known self-induced adversarial intensities $\{ \alpha_k \}$, DTS does not require explicit knowledge of the true test-time perturbation $\epsilon_t \notin \{ \alpha_1, \dots, \alpha_K \}$. Instead, DTS relies on online estimates of the value distribution shift $d_k^{(t)}$ as a proxy for robustness. As long as one policy in $\mathcal{P}$ exhibits relatively low Wasserstein shift under the unseen attack (i.e., approximates the behavior needed to withstand $\epsilon_t$), its reward proxy $r_k^{(t)} = d_{\min}^{(t)} / d_k^{(t)}$ will dominate over time. Due to posterior concentration, DTS will asymptotically prefer this policy, allowing the agent to adapt dynamically to new adversarial conditions without retraining. This property ensures that the agent is not merely robust, but \textit{adaptive} to distributional shifts caused by novel adversarial strategies.
\end{remarknn}
\begin{remarknn}
	The theoretical analysis in this work is grounded on several key assumptions, notably the Lipschitz continuity of the reward function under the value distribution shifts, and slowly-varying nature of self-induced adversarial perturbations to obtain the DTS policy. These assumptions, which are standard in the literature for achieving robustness  \cite{song2023lipsnet} , support in obtaining piecewise stationary DTS policy and also analyzing regret during test-time attacks. These assumptions are valid as they are based on slow reward change due adversarial observations which are self-induced. The assumptions are used merely to learn the adaptation, so it can deal with persistent spoofing threats which are OOD in nature.
\end{remarknn}
As per Theorem \ref{th:1}, DTS enables online selection of the policy that minimizes the distributional shift of the value function under unknown, time-varying, and even unseen adversarial attacks. Now we link the analysis, to how it makes the overall system antifragile.

\subsection{Antifragile Behaviour}
\begin{definition} \label{af_definition}
	A RL system is said to exhibit antifragility under self-induced adversarial observation perturbations if, in response to increased perturbation intensity or out-of-distribution (OOD) attacks, the system adaptively improves its expected performance (reward) over time by leveraging the variability to refine its policy selection mechanism.
\end{definition}
Now we based on the above definition, we move on to the theorem where we prove that the policy improvement under DTS makes the whole RL antifragile against unseen attacks.

\begin{theorem}[Antifragile Policy Improvement] \label{th2:antifragility}
	Let $\mathcal{P} = \{ \pi_{\alpha_1}, \dots, \pi_{\alpha_K} \}$ be a finite set of robust policies trained under adversarial intensities $\alpha_k \in [0, 1]$. Let $\kappa_t$ denote the policy selected at time $t$ by the DTS-based switching mechanism under an adversarial perturbation $\epsilon_t \notin \{\alpha_1, \dots, \alpha_K\}$, i.e., an \emph{unseen attack}. We consider the value shift satisfies the Lipschitz robustness assumption as in (\ref{eq:25}), with the reward proxy defined in (\ref{eq:26}) and DTS maintains a discounted Beta posterior as in (\ref{eq:24}). Then, under slowly evolving or piecewise-constant self-induced perturbation intensity $\epsilon_t$, the expected reward of the selected policy improves over time:
	\begin{equation}\label{eq:38}
		\lim_{t \to \infty} \frac{d}{dt} \mathbb{E}[R(\pi_{\kappa_t}; \Phi_t^{\text{adv}})] > 0.
	\end{equation}
	Thus, the system exhibits \emph{antifragile behavior} by improving performance in response to previously unseen adversarial conditions.
\end{theorem}
\begin{IEEEproof}
The proof is provided in Appendix \ref{AF:proof}.
\end{IEEEproof}
Even though $\epsilon_t$ was not seen during training, the DTS mechanism adapts using value shift observations alone. Thus, it can identify and exploit the best available robust policy without retraining. This ability to \emph{improve under perturbation} characterizes antifragility.  Detailed algorithm for DTS algorithm is provided in Appendix \ref{alg:dTS}
.\section{Results and Discussions} \label{results}
\subsection{Numerical Implementation}
The software implementation of the DDPG algorithm, used for the deconfliction of UAV navigation in the environment consisting of dynamic 3D obstacles, has been considered with respect to existing software implementation \cite{IFDSCA}. During the training phase, the initial position of the UAV was randomly considered with a mean around $\left ( 0,2,5 \right )$ and a variance between 0 and 1, while the target destination was fixed at $\left ( 10,10,5.5 \right )$. Kinematic constraints, including a maximum ascent angle of $5 \pi \slash 9$ and a maximum descent angle of $ -15 \pi \slash 36 $, were applied when updating the UAV position. In addition, a protection radius of 1.5 units was enforced around the UAV. The fixed trajectory of dynamic obstacles used for the navigation changes randomly, after every episode, so that the UAV policy is trained to deconflict against wide variety of obstacles. The reward function calculates the conflict penalty by augmenting the UAV's radius with an additional buffer of 0.4 units. The vanilla agent and the adversarial action robust models in the ensemble were trained considering a single dynamic obstacle. The action robust RL model was trained according to the Algorithm \ref{Robust_DDPG}, using the parameters specified in Table \ref{table_example}. 

In obtaining the DTS for switching between robust policies, the sampler is initially trained with a fixed value of $p_{\textup{true}}$ over 800 time steps. Subsequently, $p_{\textup{true}}$ is updated by increasing $\epsilon$ by 0.5, as shown in Figure \ref{figure4_schematic}. The discounting parameter for updating the posterior Beta distribution is considered to be 0.8. 
\subsection{Training Results}
\begin{figure}[thpb]
	\centering
	\includegraphics[scale=0.18]{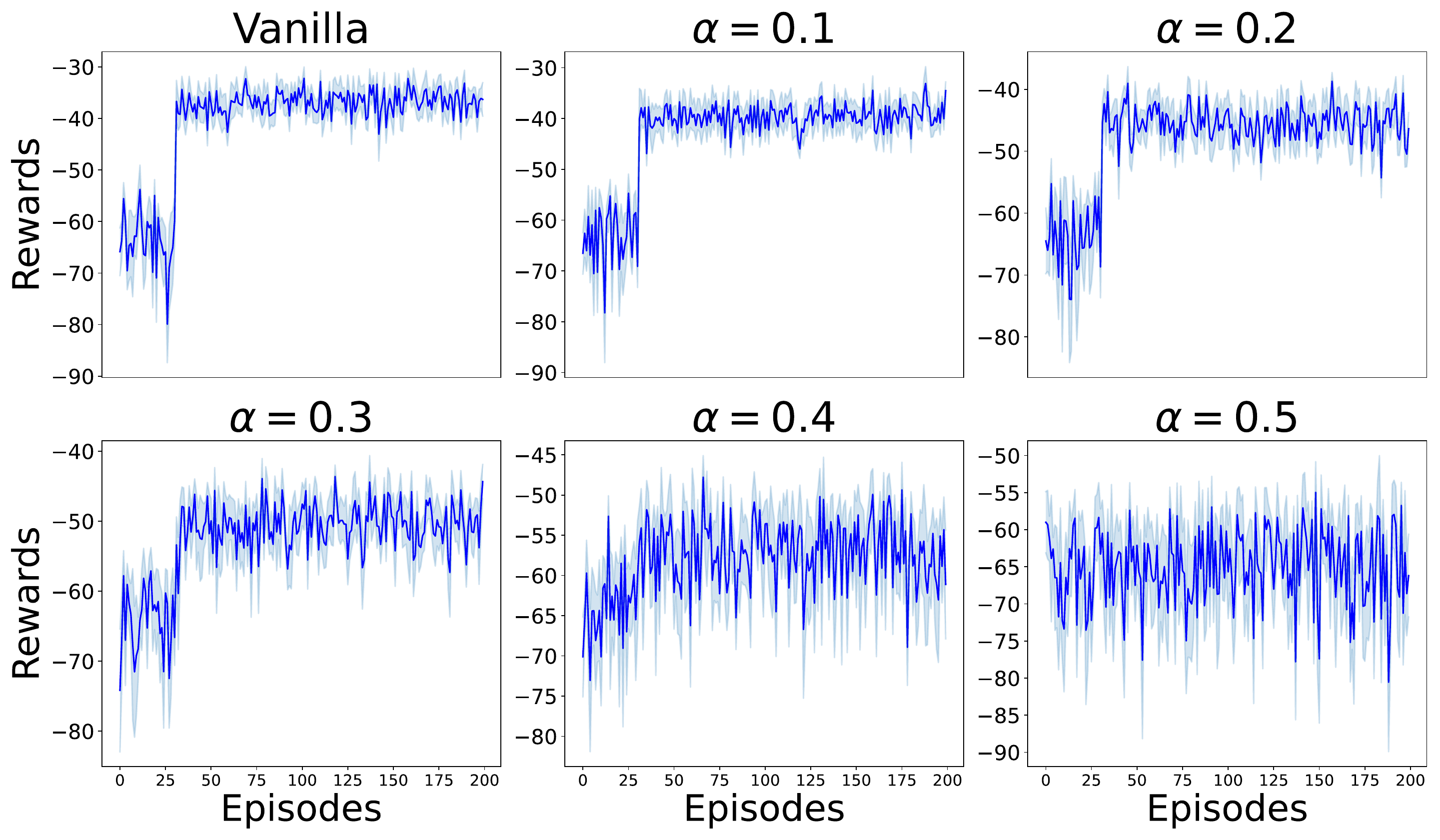}
	\caption{Convergence plot for vanilla DDPG and adversarial action robust DDPG for $\alpha = \{0.1, 0.2, 0.3, 0.4, 0.5 \}$.}
	\label{figure5_all_rewards}
\end{figure}
\begin{figure}[thpb]
	\centering
	\includegraphics[scale=0.18]{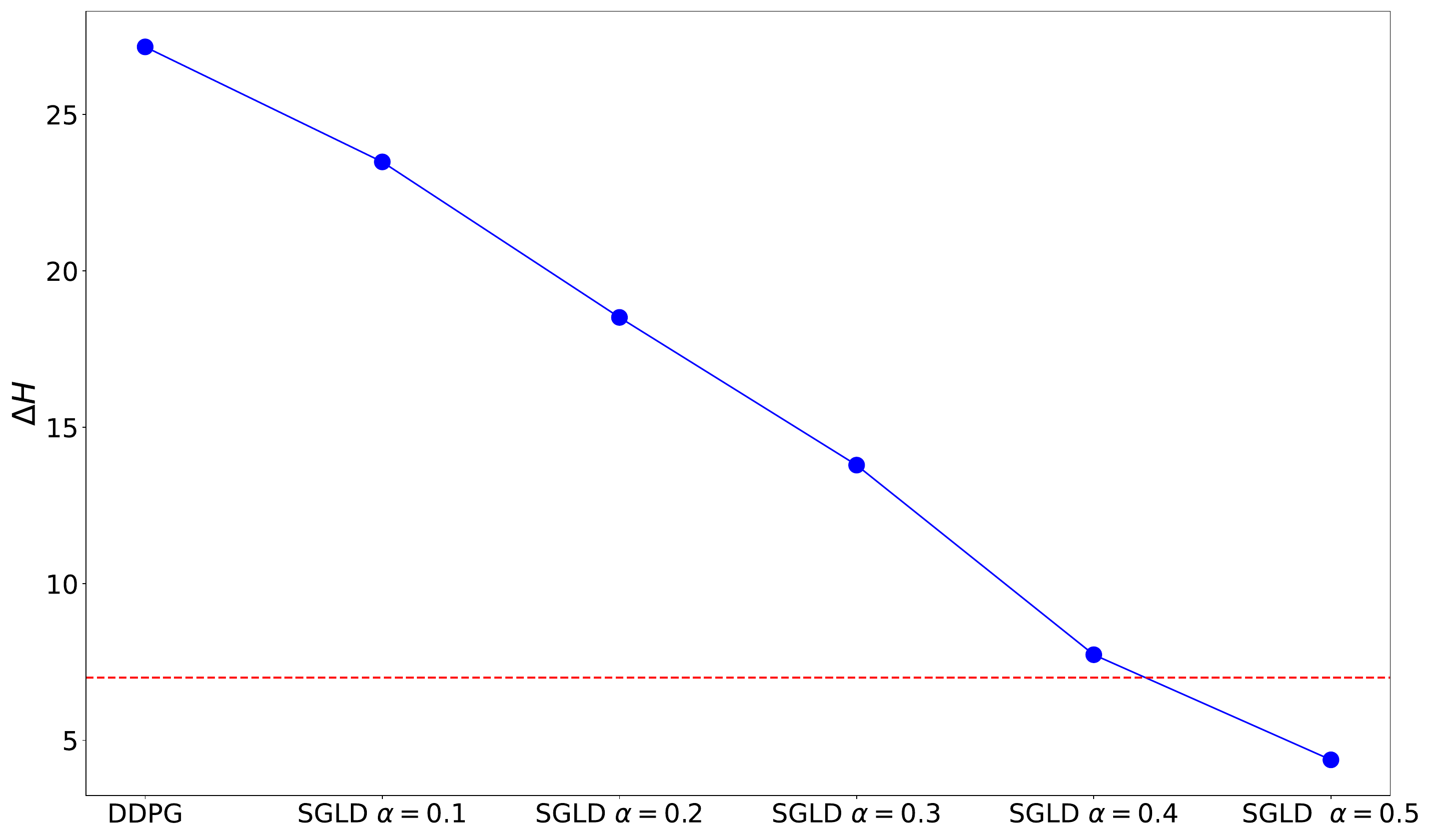}
	\caption{The difference in the entropy obtained from the rewards obtained from exploration and exploitation phase for RL training for vanilla DDPG and adversarial action robust DDPG for various $\alpha$.}
	\label{figure6_entropy_rewards}
\end{figure}

The training results, based on the cumulative reward for the episode, are illustrated in Figure \ref{figure5_all_rewards}. The action robust RL model for adversarial actions was trained using varying proportions of adversarial influence in the action space, denoted as $\alpha = \left\{ 0.1, 0.2, 0.3, 0.4, 0.5 \right\}$. As shown in Figure \ref{figure5_all_rewards}, increasing $\alpha$ leads to greater uncertainty in the converged rewards, as reflected in the widening uncertainty bands. This trend is further confirmed by the mean entropy difference $\Delta \mathbf{H}$ in Figure \ref{figure6_entropy_rewards}, where the gap between exploration and exploitation rewards decreases linearly with increasing $\alpha$. This is due to flattening of the loss landscape, due to greater randomness introduced in the policy space by the adversarial network, which in turn increased the uncertainty in the final rewards.  For $\alpha = 0.5$, there is no significant statistical difference between the rewards obtained during the exploration and exploitation phases. Consequently, robust policies corresponding to $\alpha = 0.5$ are excluded from the set of policies considered for switching during the adaptation phase.  
\subsection{Distribution Shift against Adversarial Attacks}
\begin{figure}[thpb]
	\centering
	\includegraphics[scale=0.18]{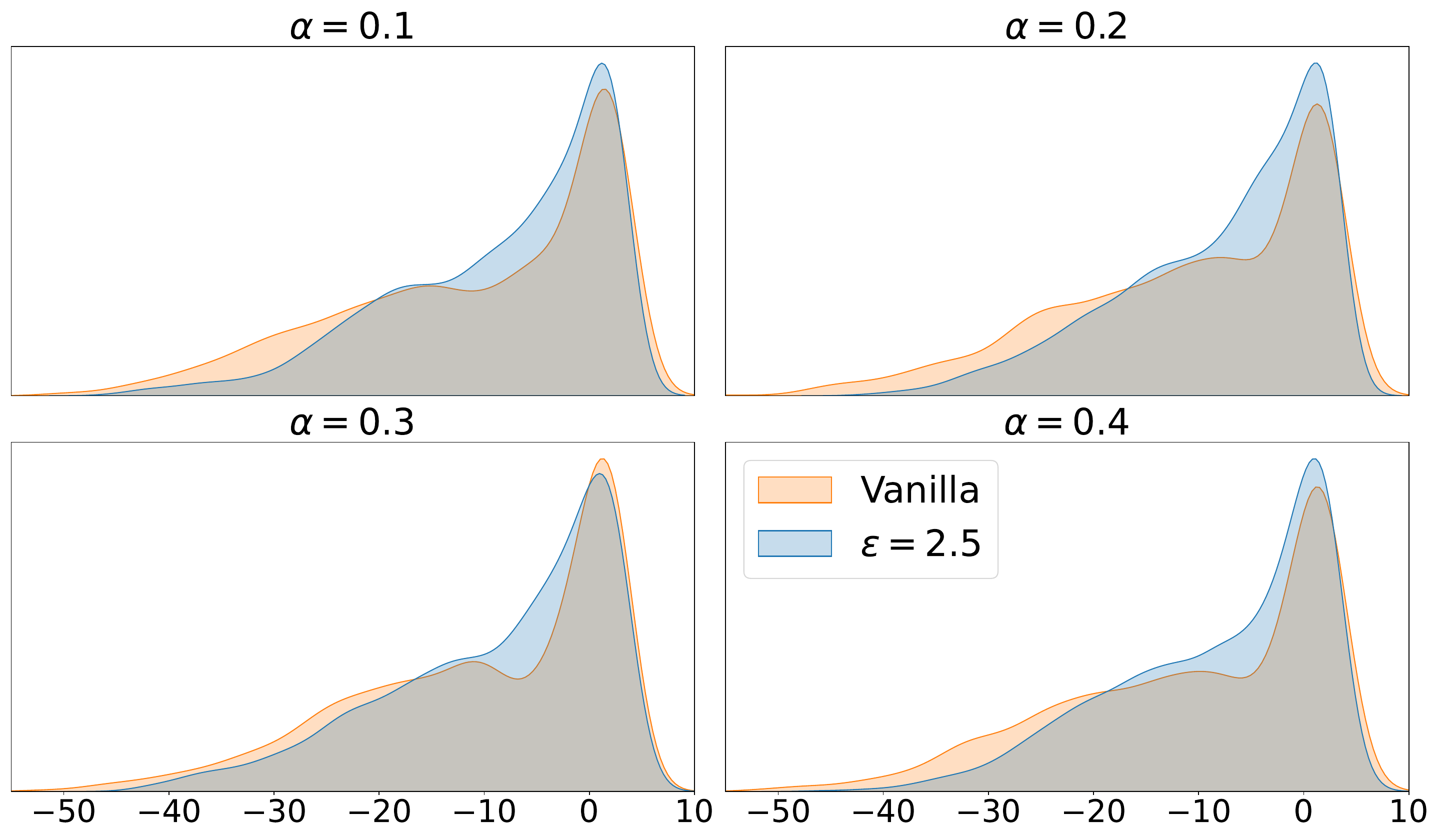}
	\caption{The distribution shift for the critic values for the action robust RL with $\alpha = \{0.1, 0.2, 0.3, 0.4 \}$ under normal operations and self-induced Frank Wolfe adversarial attack with $\epsilon = 2.5.$}
	\label{figure7_dist_shift}
\end{figure}

\begin{figure}[thpb]
	\centering
	\includegraphics[scale=0.18]{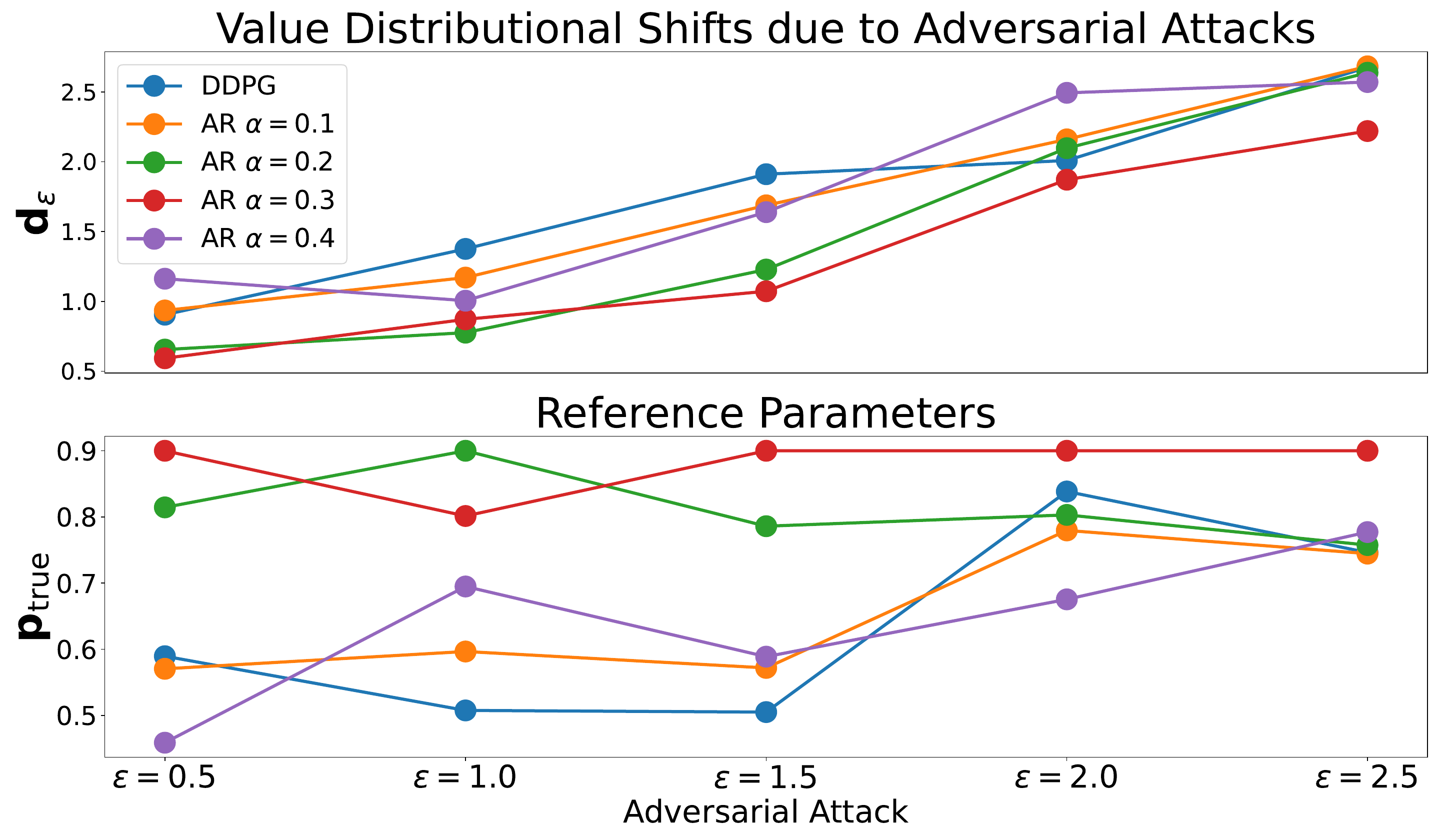}
	\caption{ a) The Wasserstein distance comparison between the critic values of vanilla DDPG and various action-robust DDPG formulations under incremental self-induced adversarial attacks.
	(b) The distributional shift induced by self-induced adversarial attacks, represented through the parameters of the Bernoulli distribution used for sampling. }
	\label{figure8_shift_plot}
\end{figure}

Once robust action policies are established with $\alpha = \left\{ 0.1, 0.2, 0.3, 0.4 \right\}$, we analyze how the state action value distribution of these robust models, along with the standard DDPG model, responds to adversarial perturbations. Adversarial observations are generated using the Frank-Wolfe optimization algorithm as stated in Algorithm \ref{alg:adv_state}, and the resulting value distributional shifts are depicted in Figure \ref{figure7_dist_shift}. Specifically, we compare value distributions under both benign conditions (no adversarial attack) and adversarial conditions where the attack magnitude is set to $\epsilon = 2.5$.
While a strict Lipschitz condition may not hold globally across all adversarial inputs, our empirical results provide evidence that small Wasserstein shifts in value distribution are correlated with the bounded changes in observed returns. In Figure \ref{figure8_shift_plot}b , we observe a monotonic, approximately linear relationship between the adversarial magnitude $\epsilon$ and the value distribution shift across the policies in the ensemble $\mathbf{d}^{\alpha}_{\epsilon}$ which suggests near-Lipschitz behaviour within the tested $\epsilon \in \left [0.5, 2.5 \right ]$. 
In particular, for $\alpha = 0.3$, the magnitude of the distributional shift remains lower compared to other robust models. This trend is evident in Figure \ref{figure7_dist_shift}, where the mode of distribution of $\alpha = 0.3$ is closely aligned with that of the standard DDPG model, in contrast to other robust policies.

\begin{figure}[thpb]
	\centering
	\includegraphics[scale=0.18]{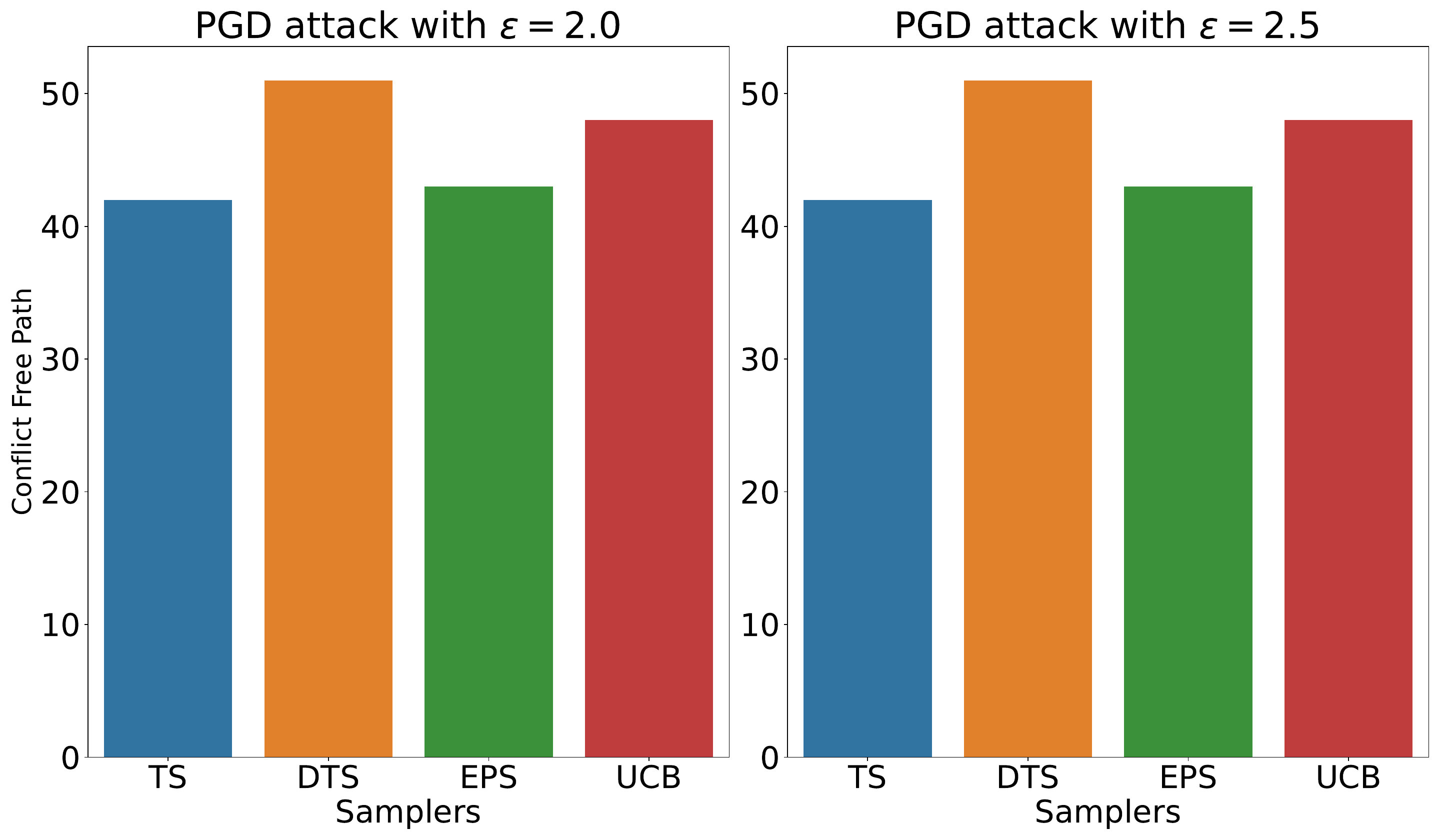}
	\caption{The number of conflict-free paths generated by the meta-policy switching algorithm over 100 episodes while comparing different sampling strategies. PGD adversarial attacks were applied at (a) $\epsilon = 2.0$ (b) $\epsilon = 2.5$. }
	\label{figure9_samp_plot}
\end{figure}

Next, we examine the behavior of $\mathbf{p}_{\textup{true}}$ that normalizes the distribution shift $\mathbf{d}_{\epsilon}$. The multiplicative factor $k_{\textup{mult}}$ that ensures a probability of 10\% of selecting non-optimal models during deployment. A distinct variation in $p_{\textup{true}}$ is observed between low and high values of $\epsilon$. Specifically, for $\alpha = 0.3$, $p_{\textup{true}}$ is the highest compared to DDPG and other robust action models. However, at $\epsilon = 1.0$, $p_{\textup{true}}$ is maximized for $\alpha = 0.2$,  reinforcing the rationale behind employing DTS sampler instead of standard TS. A sharp transition in Bernoulli rewards is also expected between $\epsilon = 0.5$ and $\epsilon = 1.0$. Furthermore, as $\epsilon$ increases from 1.5 to 2.5, the relative decline in $p_{\textup{true}}$ amongst the action robust models diminishes, with $\alpha = 0.3$ consistently yielding the highest value of $p_{\textup{true}}$. These findings suggest that the meta-policy based DTS is to be trained to switch amongst the action robust models at a higher frequency to mitigate the impact of adversarially induced distributional shifts.
Consequently, we progressively train the DTS sampler using the $p_{\textup{true}}$ values that correspond to varying $\epsilon \in \left [0.5, 2.5 \right ] $, with a critical shift at $\epsilon = 1.0$, as illustrated in Figure \ref{figure4_schematic}. Additionally, for ablation studies, alternative sampling strategies such as undiscounted TS \cite{agrawal2012analysis}, epsilon-greedy (EPS) \cite{dann2022guarantees}, and upper confidence bound (UCB) \cite{garivier2011upper} have been evaluated. Although no significant differences are observed in overall rewards, the DTS algorithm exhibits superior safety performance in testing environments containing multiple dynamic obstacles, as illustrated in Figure \ref{figure9_samp_plot} for $\epsilon = \left \{2.0, 2.5 \right\} $.
\subsection{Results against PGD Adversarial Attack}
\begin{figure}[thpb]
	\centering
	\includegraphics[scale=0.18]{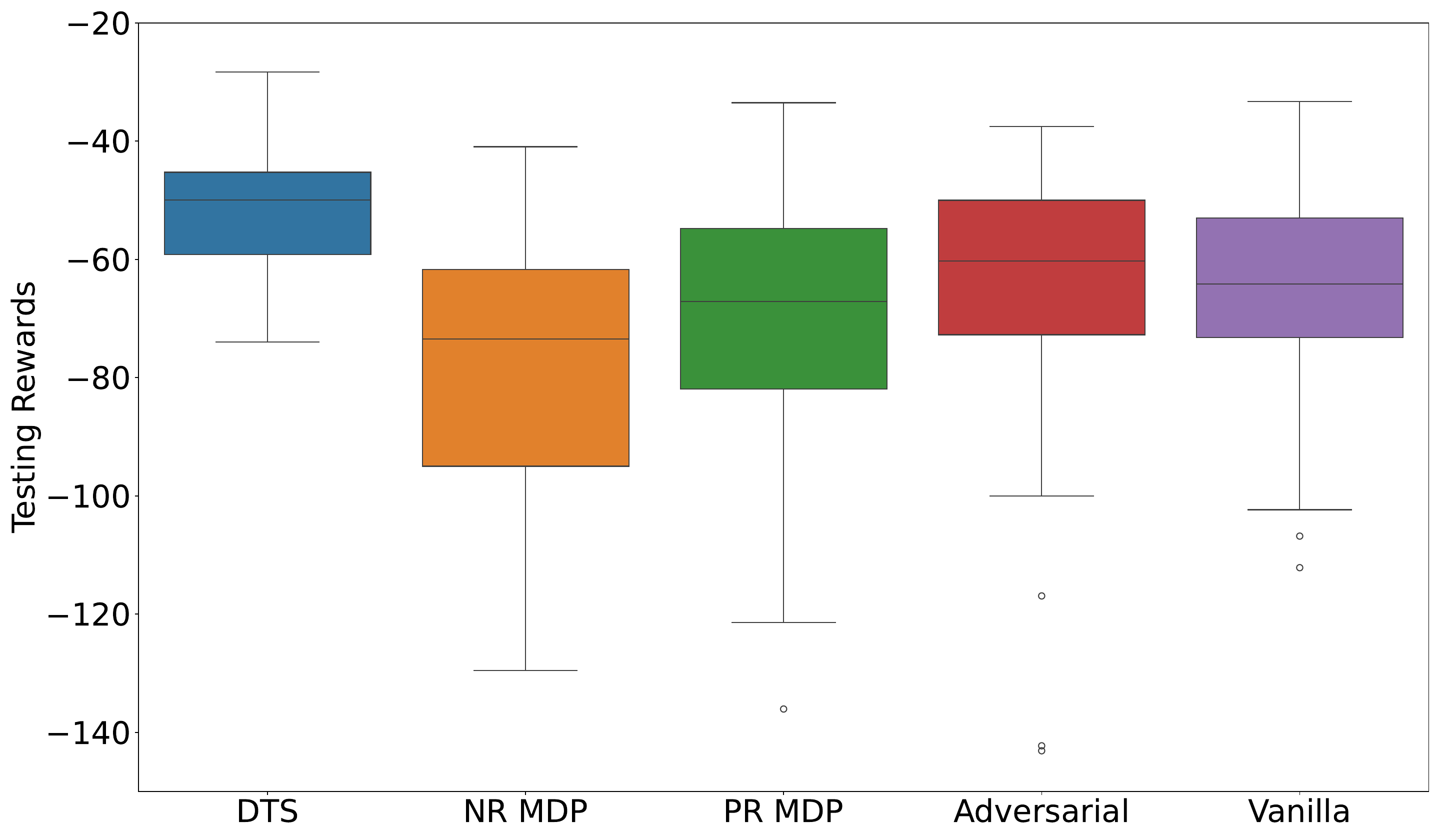}
	\caption{The rewards obtained over 100 testing episodes in a multi-obstacle environment when evaluating the DTS-based meta-policy switching algorithm against PGD adversarial attacks, in comparison to other robust RL techniques and vanilla DDPG.}
	\label{figure10_testing_rewards}
\end{figure}

We further examine navigation performance under adversarial attacks, specifically considering a Projected Gradient Descent (PGD) attack with $N_{\textup{steps}} = 50$.  The practicality of the attacks are described in Appendix \ref{test-time_threat}.
The resulting test rewards, depicted in Figure \ref{figure10_testing_rewards}, correspond to the initial randomized conditions between episodes.

\begin{figure}[thpb]
	\centering
	\includegraphics[scale=0.18]{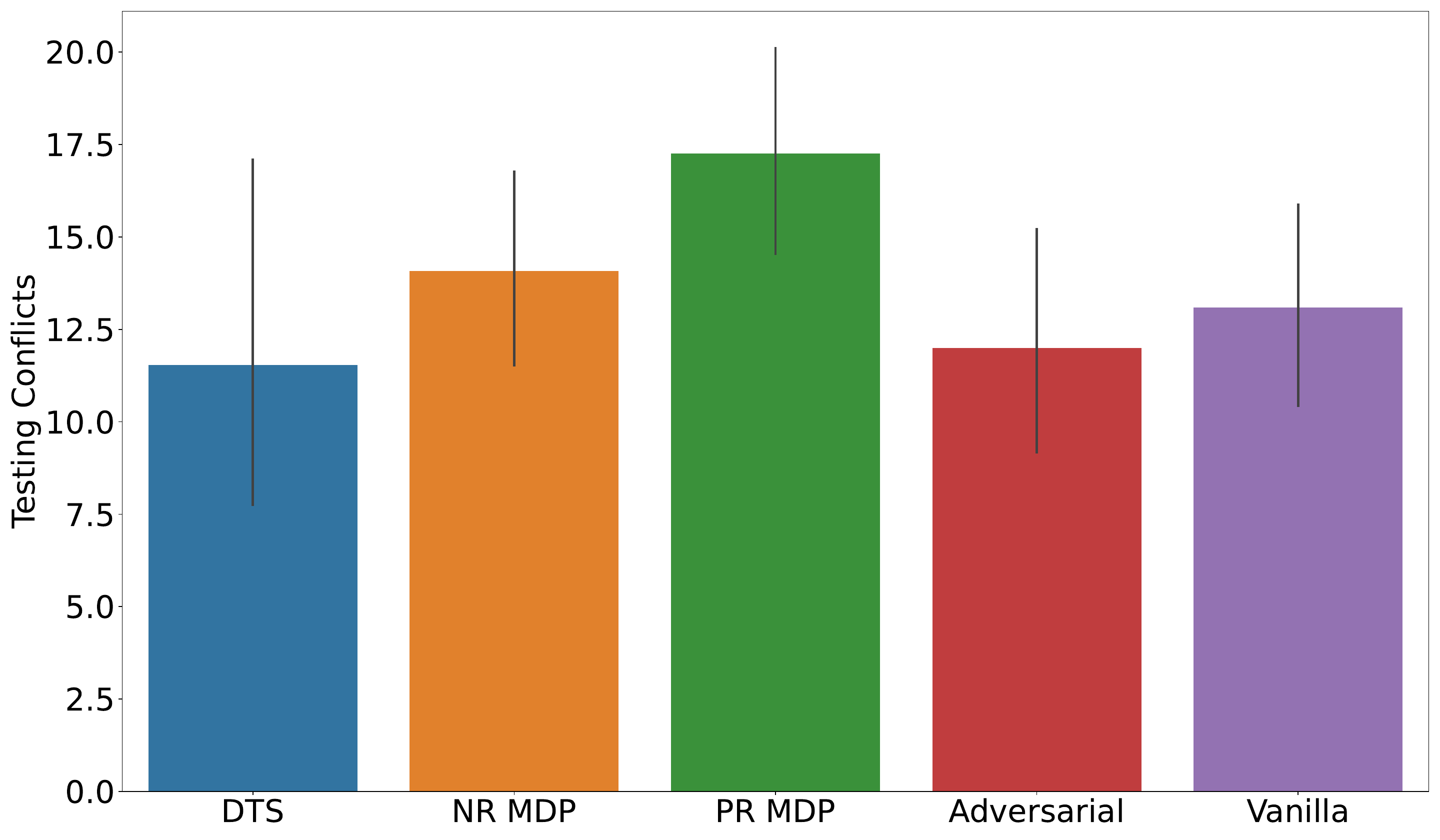}
	\caption{The total number of conflicts encountered during UAV navigation under PGD adversarial attacks in a multi-obstacle environment, comparing the DTS-based meta-policy algorithm with other robust RL techniques and vanilla DDPG over 100 episodes.}
	\label{figure11_testing_conflicts}
\end{figure}

\begin{figure}[thpb]
	\centering
	\includegraphics[scale=0.18]{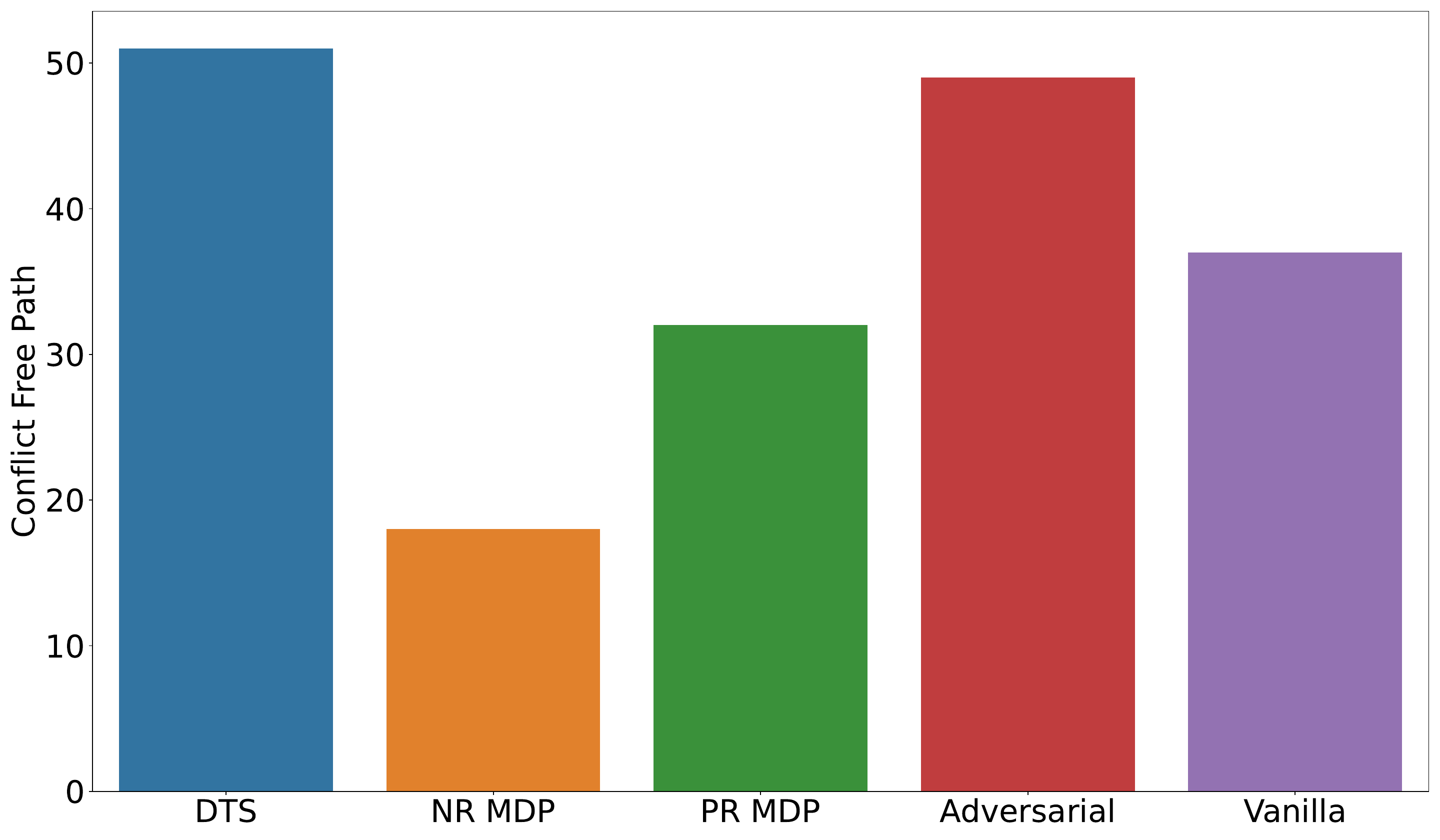}
	\caption{The total number of conflict-free paths generated during UAV navigation under PGD adversarial attacks in a multi-obstacle environment, comparing the DTS-based meta-policy switching algorithm with other robust RL techniques and vanilla DDPG over 100 episodes.}
	\label{figure12_zero_conflicts}
\end{figure}

The meta-policy switching algorithm  demonstrates superior performance, achieving higher testing rewards compared to the baseline algorithms. Furthermore, the variance in rewards is lower for the DTS-based approach, indicating increased stability. Safety performance is assessed by analyzing both the total number of conflicts and the proportion of conflict-free trajectories in the testing environment in Figures \ref{figure11_testing_conflicts} and \ref{figure12_zero_conflicts}. The DTS algorithm significantly outperforms the robust and vanilla standard RL models in minimizing conflicts. However, while DTS marginally surpasses adversarial learning methods in overall performance, the latter exhibits a lower variance in conflict occurrences. Despite this, Figure \ref{figure10_testing_rewards} reveals that adversarial learning achieves lower overall rewards than the DTS-based method. This suggests that while adversarial learning improves collision safety, it does so at the expense of efficient path planning, thereby increasing vulnerability to alternative forms of attack.
\subsection{Results against Spoofing Attacks for Hijacking}
\begin{figure}[thpb]
	\centering
	\includegraphics[scale=0.18]{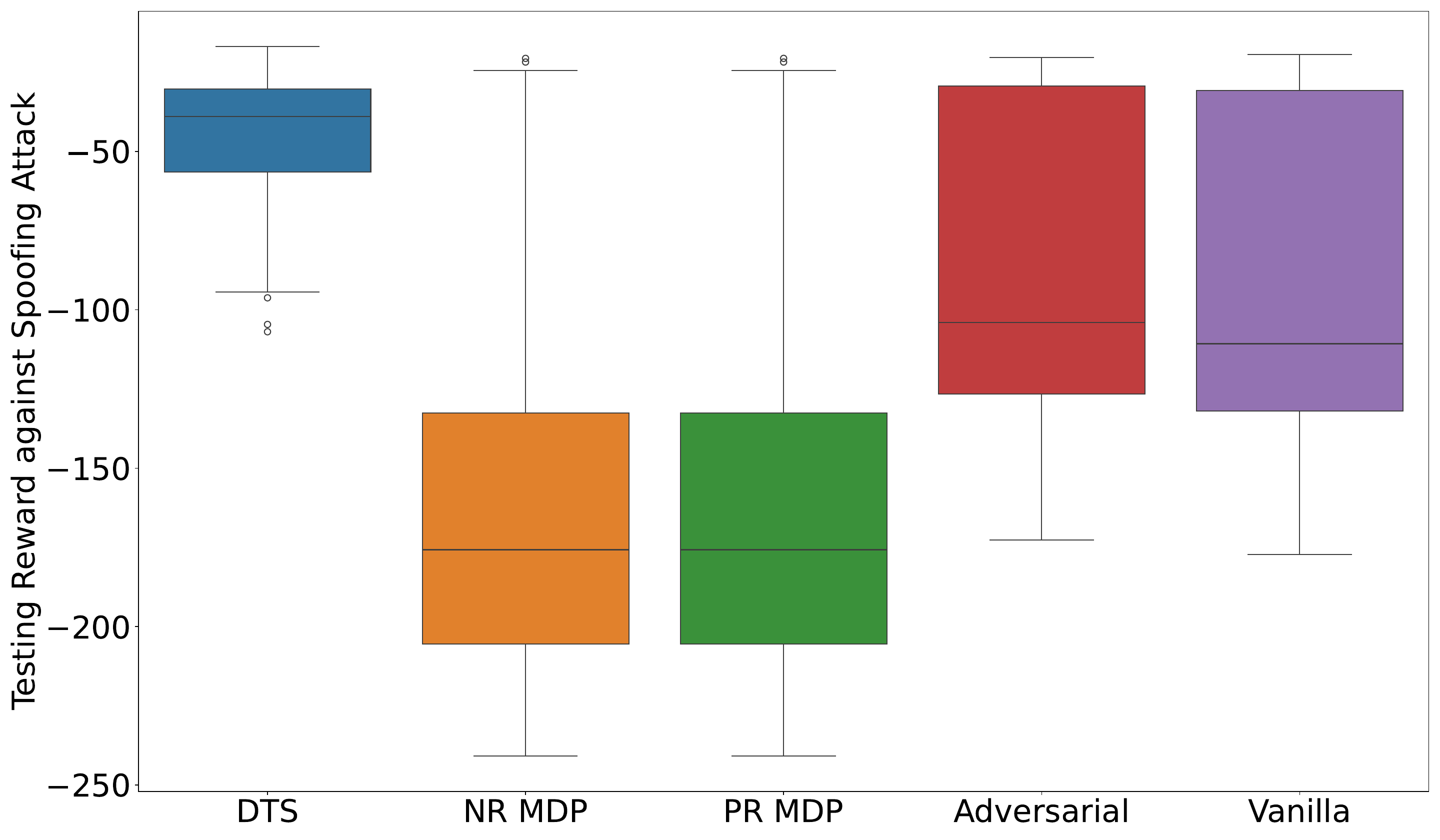}
	\caption{The rewards obtained over 100 testing episodes when evaluating the antifragile algorithm with a DTS sampler against other robust RL techniques and vanilla DDPG under high out-of-distribution attacks like GNSS spoofing.}
	\label{figure13_spoofing_rewards}
\end{figure}

A separate analysis considers a different adversarial strategy, in which the position of the UAV is spoofed to facilitate hijacking as detailed in Appendix \ref{test-time_threat}. This attack is commonly used to manipulate position inputs, forcing UAVs to elongate flight paths through specific airspace zones, increasing exposure to hijacking attempts. To simulate this, the position of the UAV in each time step is perturbed with a uniform random bias $\mathcal{U} \left (0.04, 0.06 \right )$. The objective is to determine whether the UAV deviates significantly from its intended trajectory. 
As illustrated in Figure \ref{figure13_spoofing_rewards}, the DTS-based meta-policy algorithm maintains an optimal trajectory with minimal deviation, unlike alternative methods that exhibit excessive compensation in response to adversarial perturbations. Specifically, the rewards for NR-MDP and PR-MDP are notably lower due to their reliance on fixed countermeasures. In contrast, the adaptive nature of the DTS sampler, trained in non-stationary Bernoulli rewards, enables effective path planning while mitigating deviations induced by spoofing attacks. Consequently, the proposed approach ensures improved resilience against hijacking attempts while preserving efficient navigation.
Hence, we can conclude that the test rewards degrade smoothly with out-of-distribution adversarial attacks under non-switching baselines, while DTS mechanism preserves performance indicating that the degradation does not exhibit chaotic discontinuities in practice. The use of testing attacks as explained in Section\ref{test:attacks}, is a security-stress test where our goal is to measure the adaptability of the system to different types of OOD perturbations. In real deployments, even partial adversarial knowledge can lead to policy appromixation attacks, control channel spoofing or physical-space decoy placement which is effectively abstracted from our synthetic adversarial observations. 
\section{Conclusions} \label{concl}
In this study, a meta-policy adaptive framework has been developed, employing a mechanism for switching between robust policies to enhance adaptability in the presence of self-induced adversarial observation attacks. DTS is utilized to optimally transition between robust models, mitigating the overall distributional shift in value estimation under adversarial perturbations. The theoretical framework is derived which states the optimality of DTS based switching and inducing antifragile behaviour against unseen adversarial attacks.

The proposed DTS-based switching approach is evaluated against progressively stronger adversarial attacks, including hijacking-induced spoofing, and benchmarked against conventional robust and standard RL techniques. The key findings of this study are as follows.
\begin{itemize}
	\item \textbf{Non-stationarity of action robust policies under adversarial attacks:} Analyzing various robust models under escalating adversarial attacks reveals that a single robust model does not maintain optimal performance across different attack intensities. This underscores the necessity of dynamic policy switching to achieve optimal resilience. The ablation studies with respect to other bandit algorithms show that DTS maintains the best safety as compared to other algorithms.
	
	\item \textbf{Enhanced Robustness Against Strong PGD Adversarial Attacks:} The DTS-based meta-policy switching framework outperforms standard and benchmark robust RL methods in countering stronger Projected Gradient Descent (PGD) attacks, particularly in terms of minimizing navigation path lengths with average reward improvement of 50\% and increased average number of conflict-free trajectories by around 5\% with respect to adversarial RL and 10\% with respect to probabilistic robust RL.
	
	\item \textbf{Navigation Resilience to Spoofing Attacks:} The meta-policy adaptive switching framework demonstrates superior resilience to previously unseen spoofing attacks compared to robust and standard RL models. This highlights the core principle of antifragility, in which a system trained under weaker adversarial conditions exhibits the capacity to counter unforeseen attack scenarios effectively.
	
\end{itemize}

In real-world scenarios involving highly dynamic or abruptly changing adversarial behaviors, these assumptions behind learning meta-policy switching may not hold. In such cases, additional mechanisms—such as online change-point detection \cite{panda2025real} or reset-free meta-learning \cite{lu2020reset} —should be integrated to maintain antifragility. Developing such extensions constitutes a key direction for future work.

\printbibliography

\appendices
\renewcommand\thesubsection{\thesection.\Roman{subsection}}
\counterwithin{equation}{section}
\newpage
\section{Interfered Flow Field} \label{inter_flow}
The interfered flow field is the modification of the flow field that affects the initial flow field due to the presence of obstacles. The 3D obstacle is designed as a standard convex polyhedron, as:
\begin{equation} \label{eq:A1}
	\Gamma\left ( \mathbf{P} \right ) = \left (  \frac{x-x_0}{\breve{a}}\right )^{2\breve{p}} + \left (  \frac{y-y_0}{\breve{b}}\right )^{2\breve{q}} + \left (  \frac{z-z_0}{\breve{c}}\right )^{2\breve{r}}. 
\end{equation}
In this case, $\breve{p}$, $\breve{q}$, $\breve{r}$, $\breve{a}$, $\breve{b}$, $\breve{c}$ determine the shape of the convex polyhedron of the obstacle, and $\left ( x_0, y_0, z_0 \right )$ determine centre of the convex polyhedron. If we consider $N$ obstacles in the space, the interfered flow field is represented by a disturbance matrix as: 
\begin{equation} \label{eq:A2}
	\overline{M}\left ( \mathbf{P} \right ) = \sum_{n=1}^{N} \mathfrak{w}_n \left ( \mathbf{P} \right ) M_n \left ( P \right ),
\end{equation}
where $\mathfrak{w}$ is the disturbance weighting factor for the $N$ obstacles defined as: 
\begin{equation} \label{eq:A3}
	\mathfrak{w}_n \left ( \mathbf{P} \right ) = \left\{\begin{matrix}
		1 \; \; \; \;  N = 1, \\
		\prod_{i=1, i \neq n}^{N} \frac{\Gamma_i \left ( P \right ) - 1}{\Gamma_i \left ( P \right ) - 1 + \Gamma_n \left ( P \right ) - 1}.
	\end{matrix}\right. 
\end{equation}
Here we define the radial normal vector along the 3D obstacle $\mathbf{t}_n$ as:
\begin{equation} \label{eq:A4}
	t_n \left ( \mathbf{P} \right ) = \begin{bmatrix}
		\frac{\partial \Gamma_n \left ( \mathbf{P} \right )}{\partial x} & \frac{\partial \Gamma_n \left ( \mathbf{P} \right )}{\partial y} & \frac{\partial \Gamma_n \left ( \mathbf{P} \right )}{\partial z}  \\
	\end{bmatrix}.
\end{equation}
We consider two vectors $h_{n,1} \left ( \mathbf{P} \right )$, $h_{n,2} \left ( \mathbf{P} \right )$ which are orthogonal to the radial vector $\mathbf{t}_n$ and also mutually orthogonal to each other, given as:
\begin{equation} \label{eq:A5}
	\begin{gathered}
		h_{n,1} \left ( \mathbf{P} \right ) = 
		\begin{bmatrix}
			\frac{\partial \Gamma_n \left ( \mathbf{P} \right )}{\partial y} \\  
			-\frac{\partial \Gamma_n \left ( \mathbf{P} \right )}{\partial x}\\
			0  \\
		\end{bmatrix}^T, \\
		h_{n,2} \left ( \mathbf{P} \right ) = \begin{bmatrix}
			\frac{\partial \Gamma_n \left ( \mathbf{P} \right )}{\partial x}  \cdot \frac{\partial \Gamma_n \left ( \mathbf{P} \right )}{\partial z}  \\  
			\frac{\partial \Gamma_n \left ( \mathbf{P} \right )}{\partial x}  \cdot \frac{\partial \Gamma_n \left ( \mathbf{P} \right )}{\partial z}  \\
			-\left ( \frac{\partial \Gamma_n \left ( \mathbf{P} \right )}{\partial x}  \right )^2 -\left ( \frac{\partial \Gamma_n \left ( \mathbf{P} \right )}{\partial y}  \right )^2  \\
		\end{bmatrix}^T.
	\end{gathered}
\end{equation}
The plane $S$ formed by $h_{n,1}  \left ( \mathbf{P} \right )$, $h_{n,2}  \left ( \mathbf{P} \right )$ is perpendicular to $t_n$. Hence, we can write any unit vector on the plane $S$ as, 
\begin{equation} \label{eq:A6}
	\begin{bmatrix}
		\cos  \vartheta_n & \sin \vartheta_n & 0  \\
	\end{bmatrix},
\end{equation}
where $\vartheta_n$ is the direction coefficient affecting the tangential disturbance. The disturbance matrix while considering a single obstacle is given as:
\begin{equation} \label{eq:A7}
	\begin{gathered}
		\mathbf{M}_n \left ( \mathbf{P} \right ) = \mathbf{M}_{n1} \left ( \mathbf{P} \right )
		+ \mathbf{M}_{n2} \left ( \mathbf{P} \right ),  \\
		\mathbf{M}_n = \mathbf{I} - \frac{\mathbf{t}_n \left ( \mathbf{P} \right ) \mathbf{t}_n \left ( \mathbf{P} \right )^T}{\left| \Gamma_n \left ( \mathbf{P} \right ) \right|^{\frac{1}{\varrho_n}}\mathbf{t}_n \left ( \mathbf{P} \right )^T \mathbf{t}_n \left ( \mathbf{P} \right )} + \frac{h_n \left ( \mathbf{P} \right ) \mathbf{t}_n \left ( \mathbf{P} \right )^T}{\left| \Gamma_n \left ( \mathbf{P} \right ) \right|^{\frac{1}{\varsigma_n}}h_n \left ( \mathbf{P} \right )^T \mathbf{t}_n \left ( \mathbf{P} \right )}.
	\end{gathered}
\end{equation}
Here we can define $\varrho_n$ and $\varsigma_n$ using the given formulas,
\begin{equation} \label{eq:A8}
	\left\{\begin{matrix}
		\varrho_n = \varrho_0 \cdot \exp \left ( 1 - \frac{1}{\bar{d}\left ( P, P_d \right ) \bar{d}\left ( P, O_n \right )}  \right ),\\
		\varsigma_n = \varsigma_0 \cdot \exp \left ( 1 - \frac{1}{\bar{d}\left ( P, P_d \right ) \bar{d}\left ( P, O_n \right )}  \right ).
	\end{matrix}\right.
\end{equation}
Here, $d\left ( \mathbf{P}, O_n \right )$ represents the distance between the end of the path and the surface of the $N$-th obstacle, $\varrho_0$,$\varsigma_0$ represents the response coefficient of the obstacle. The dynamic obstacle speed threat is represented as
\begin{equation} \label{eq:A9}
	v_{\textup{obs}} \left ( \mathbf{P} \right ) = \sum_{n=1}^{N} \mathfrak{w}_n \left ( \mathbf{P} \right ) \exp \left [ \frac{-\Gamma_n \left ( \mathbf{P} \right )}{\Upsilon} \right ] V_{\textup{obs}}^{n} \left ( \mathbf{P} \right ).
\end{equation}
where $V_n^{\textup{obs}}$ is considered as the moving speed of the $N$th obstacle, $\Upsilon$ is a constant greater than zero. 
\section{Dual Threat Framework} \label{test-time_threat}
We elaborate on the threat realism and relevance of each class below.
\begin{itemize}
	\item \textbf{Observation-Space Attacks via Policy Inference (Grey-Box Threat Realism):} These attacks emulate adversarial manipulation of the UAV’s state representation through inferred policy models, relying on the following feasible steps:
	\begin{itemize}
		\item \textbf{Step 1:} In cooperative airspace environments, UAV telemetry—comprising position, velocity, and trajectory updates—is frequently broadcast over openly accessible communication channels such as ADS-B \cite{manesh2017analysis, strohmeier2014security}. An adversary with passive eavesdropping capabilities can intercept these transmissions without detection.
		
		\item \textbf{Step 2:} Leveraging the intercepted data, the adversary reconstructs the UAV’s input-output behavior using imitation learning frameworks such as Generative Adversarial Imitation Learning (GAIL) \cite{ho2016generative} or behavior cloning \cite{morales2004learning}. This yields a surrogate policy model $\hat{\pi}$ that approximates the UAV’s original navigation policy.
		
		\item \textbf{Step 3:} Once $\hat{\pi}$ is obtained, the adversary can generate adversarial perturbations using gradient-based methods such as  PGD \cite{gupta2018cnn}, effectively launching white-box style attacks on the approximated policy. These perturbations can be transmitted back to the UAV’s ADS-B IN interface, as malicious UTM operator, thereby influencing control decisions in a stealthy and coordinated manner.
	\end{itemize}	
	The feasibility of such inference-based attacks has been demonstrated in our prior work on adversarial mobility and surveillance \cite{chu2024survey}, where unencrypted trajectory data were exploited to construct targeted adversarial strategies. Since the induced distributional shifts due to PGD attacks resemble to those introduced by Frank-Wolfe-style perturbations, we categorize such attacks as \textit{low out-of-distribution (OOD)} adversarial scenarios.
	\item \textbf{GNSS Spoofing Attacks (Black-Box Signal-Space Threats):} In contrast to observation-space attacks, GNSS spoofing directly manipulates the UAV’s positional awareness by injecting falsified satellite signals. These attacks operate at the signal layer and do not require any access to the policy network or its gradients, thus constituting a fully black-box threat model. The attacker exploits pseudorange manipulation \cite{panda2025real} to introduce consistent positional drift, misleading the UAV into computing paths based on erroneous geolocation data. Spoofing can be realized using low-cost software-defined radios (SDRs), constrained primarily by the proximity of the attacker to the UAV and the ability to overpower legitimate GNSS signals. This can lead to critical safety violations, including extended flight path, forced penetration into restricted zones, or increased collision risks. GNSS spoofing induces high-magnitude state-space shifts that are not gradient-driven and are thus semantically and statistically distinct from algorithmic perturbations from Frank-Wolfe adversarial attacks, we classify them as \textit{high OOD} adversarial threats.	
\end{itemize}
The inclusion of both adversarial observation perturbations and GNSS spoofing captures two fundamental axes of UAV system vulnerability: (1) \textit{control-layer adversaries}, where the attack targets decision-making logic via state perturbations, and (2) \textit{sensor-layer adversaries}, which compromise state estimation at the signal acquisition level. The proposed DTS-based policy switching framework enables robust adaptation across this threat spectrum, exhibiting resilience against both low and high OOD adversarial perturbations. In the following section, we present the theoretical guarantees underlying this adaptive robustness.
\section{Parameters for Training Robust Policies}
\begin{table}
	\caption{Adversarial Action Robust RL Parameters}
	\label{table_example}
	\centering
	\begin{tabular}{c||c}
		\hline
		\bfseries Parameter & \bfseries Value\\
		\hline\hline
		Exploration Episodes & 30 \\
		Replay buffer size & 1e6  \\
		Minibatch size &  128 \\
		Discount factor &  0.99 \\
		Target network soft update rate & 0.01  \\
		Agent and adversary network learning rate &  0.01  \\
		Agent and adversary network momentum & 0.9 \\
		Adversary noise parameter & 0.1 \\ 
		Weight decay & 5e-4 \\
		Critic learning rate & 1e-3   \\
		Neural network optimizer & Adam  \\
		Exploration noise & $\mathcal{N} \left ( 0.05, 0.02 \right )$  \\
		Maximum steps per episode & 500   \\
		Total number of episodes &  200 \\
		Total neural network layers &  3 \\
		Robustness update rate $\left (\beta  \right ) $ & 0.9  \\
		Number of neurons in each layer & 128  \\
		Number of dynamic obstacles in testing scenario & 4 \\
		\hline
	\end{tabular}
\end{table}
\section{Sampling Based Solution for Robust RL} \label{langevin}
Sampling-based methods are employed to derive the mixed Nash equilibrium from the objective function in (\ref{eq:19}), as outlined in \cite{hsieh2019finding}. This equilibrium is then utilized to address RL challenges \cite{kamalaruban2020robust}. In order to avoid the saddle points, isotropic gradient noise is introduced into the stochastic gradient descent (SGD) process used to minimize the loss functions during the training of agent and adversarial policy networks. By adaptively scaling this noise as per higher-order curvature measures, such as Fisher scoring, the noise can be preconditioned to achieve improved convergence properties \cite{bhardwaj2019adaptively}. When optimizing the loss function using SGD concerning the parameter $\Theta$, the update equation at each step is given by: 
\begin{equation} \label{eq:D1}
	\hat{g_s}\left ( \Theta_t \right ) \leftarrow \nabla_{\Theta} \hat{\mathcal{L}}_s \left ( \Theta_t \right ).
\end{equation}
Here,  $\hat{\mathcal{L}}_s \left ( \Theta_t \right )$ represents the stochastic estimate of the loss function. Hence, the update equation can be represented as:
\begin{equation} \label{eq:D2}
	\Theta_{t+1} \leftarrow \Theta_t - \eta \left ( \hat{g_s} \left ( \Theta_t \right ) \right ). 
\end{equation}
Incorporating Gaussian noise in the parametric update $\mathcal{N} \left ( 0, \mathfrak{e} \right )$:
\begin{equation} \label{eq:D3}
	\begin{gathered}
		\zeta_t \sim \mathcal{N} \left ( 0, \mathfrak{e}  \right ), \\
		\Theta_{t+1} \leftarrow \Theta_t - \eta \left ( \hat{g}_s \left ( \Theta_t \right )  + \zeta_t \right ).
	\end{gathered}
\end{equation}
The uniform scaling of noise can lead to improper parameter update scaling, potentially slowing the algorithm's training speed and resulting in convergence to suboptimal minima \cite{luoadaptive}. In order to issue this, adaptive preconditioners are employed, such as RMSProp \cite{zou2019sufficient}, which approximates the inverse of the second-order moments of the gradient update using a diagonal matrix. Bhardwaj et al. \cite{bhardwaj2019adaptively} propose an adaptive preconditioned noise that takes advantage of a diagonal approximation of the second-order gradient updates' moments to enhance the training efficiency of first-order methods. Consequently, noise scaling is adjusted proportionally to accelerate training as follows:
\begin{equation} \label{eq:D4}
	\begin{gathered}
		\bm{C}_t \leftarrow  \rho \bm{C}_{t-1} + \left ( 1 - \rho \right ) \left ( \hat{\bm{g}_s} \left ( \Theta_t \right ) - \bm{\mu}_t \right )\left ( \hat{\bm{g}_s} \left ( \Theta_t \right ) - \bm{\mu}_{t-1} \right ), \\ 
		\zeta_t \sim \mathcal{N} \left ( \bm{\mu}_t, \bm{C}_t \right ), \\
		\Theta_{t+1} \leftarrow \Theta_t - \eta \left ( \hat{\bm{g}}_s \left ( \Theta_t \right ) + \psi \zeta_t \right ).
	\end{gathered}
\end{equation}
In equation (\ref{eq:D4}), we observe that the noise covariance pre-conditioner, scales the noise proportionally along the dimensions with larger gradients. This approach helps to escape saddle points and navigate the policy space more effectively while dealing with broader loss minima, ultimately leading to faster convergence and improved solutions. Therefore, incorporating adaptive noise into the loss function allows optimization of both the critic loss and the agent and adversarial policy loss functions. If $\alpha_i$ as the adversarial contribution, then the general action $a_t$ is expressed as follows:
\begin{equation} \label{eq:D5}
	a_t = \alpha_i \nu_{\omega}\left ( \Phi_t \right ) + \left ( 1 - \alpha_i \right ) \pi_{\theta} \left ( \Phi_t \right ).
\end{equation}
If we consider a batch of samples $B = \left\{ \left ( \Phi_t, a_t, r_t, {\Phi}', \mathfrak{d} \right ) \right\}$ from the replay buffer $\mathcal{D}$, then we compute the critic target $y_{\textup{targ}}$  as follows:
\begin{equation} \label{eq:D6}
	y_{\textup{targ}} =  r_t  + \gamma \left ( 1 - \mathfrak{d} \right ) Q_{\phi_{\textup{targ}}} \left ( \Phi', \left ( 1 - \alpha_i \right ) \mu_{\theta} + \alpha_i \nu_{\omega} \right ).
\end{equation}
Thus target $y_{\textup{targ}}$ can be utilized to formulate the critic loss function as, 
\begin{equation} \label{eq:D7}
	L\left ( \phi \right ) = \frac{1}{N} \sum_{B = \left\{ \left ( \Phi_t, a_t, r_t, {\Phi}', \mathfrak{d} \right ) \right\}} \left ( y\left ( r, \Phi',d \right ) - Q_{\phi} \left ( \phi, a \right )  \right )^2. 
\end{equation}
We can also compute the agent and adversary policy updates as:
\begin{equation} \label{eq:D8}
	\begin{gathered}
		\nabla_{\theta} \widehat{J \left ( \theta, \omega_t \right )} = \frac{1 - \alpha_i}{N}  \sum_{\Phi \in \mathcal{D}} \nabla_{\theta} \mu_\theta \left ( \Phi \right )  \nabla_a Q_{\phi} \left ( \Phi, a \right ) ,   \\
		\nabla_{\omega} \widehat{J \left ( \theta_t, \omega \right )} = \frac{\alpha_i}{N} \sum_{\Phi \in \mathcal{D}} \nabla_{\omega} \nu_\omega \left ( \Phi \right ) \nabla_a Q_{\phi} \left ( \Phi, a \right ). 
	\end{gathered}
\end{equation}
To minimize the loss functions of critics and policy given by equations (\ref{eq:D6}) and (\ref{eq:D7}) respectively, we incorporate SGLD noise into the update equations as specified in (\ref{eq:D4}). This approach is taken to update the joint critic parameters $\phi$ and the agent and adversary policy parameters $\left\{ \theta, \omega \right\}$. If we consider $\bm{g} = \left \{ \nabla_{\theta} J, \nabla_{\omega} J   \right \} $, $\bm{\mu}_t = \left \{ \mu_{\theta}, \mu_{\omega}   \right \} $, $\bm{C}_t = \left \{ C_{\theta}, C_{\omega}   \right \}$ according to the dynamics of SGLD given in (\ref{eq:D4}), the update equation for the agent and the adversary parameter to obtain the mixed Nash equilibrium for the max-min objective in (\ref{eq:19}) for the $k^{\textup{th}}$ sample from the buffer at time $t$, is given as:
\begin{equation}\label{eq:D9}
	\theta^{k+1}_{t} \leftarrow \theta^{k}_{t} + \eta \left ( \nabla_{\theta} \widehat{J\left ( \theta, \omega_t \right )} + \psi \xi_t  \right ),
\end{equation}
\begin{equation}\label{eq:D10}
	\omega^{k+1}_{t} \leftarrow \omega^{k}_{t} - \eta \left ( \nabla_{\omega} \widehat{J\left ( \theta, \omega_t \right )}  + \psi \xi_t  \right ).
\end{equation}
Here $\psi$ represents the noise parameter. The parameters updated with the time $t$ is given by
\begin{equation}\label{eq:D11}
	\bar{\omega}_t = \left ( 1 - \beta  \right ) \bar{\omega}_t + \beta \omega^{\left (k+1  \right )}_t ;
	\; \bar{\theta}_t = \left ( 1 - \beta  \right ) \bar{\theta}_t + \beta \theta^{\left (k+1  \right )}_t.
\end{equation}
Obtain the parameters for the next time duration as,
\begin{equation}\label{eq:D12}
	\omega_{t+1} = \left ( 1 - \beta  \right ) \omega_t + \beta \bar{\omega}_t ;  \; \theta_{t+1} = \left ( 1 - \beta  \right ) \theta_t + \beta \bar{\theta}_t.
\end{equation}
The overall training algorithm to obtain an ensemble set of agent and adversarial policy networks is explained in Algorithm \ref{Robust_DDPG}.
\begin{algorithm}
	\caption{Ensemble Adversarial Action Robust RL}
	\label{Robust_DDPG}    
	Initialize: Policy parameters $\omega$, $\theta$ and $Q$ function parameter $\phi$.   \\
	Initialize: The target network parameters $\omega_{\textup{targ}} \leftarrow  \omega_1$ , $\theta_{\textup{targ}} \leftarrow  \theta_1$, and $\chi_{\textup{targ}} \leftarrow \chi$.\\
	Initialize: Replay buffer \\
	Initialize: Ensemble Policy Set $\mathcal{P} \leftarrow \emptyset$ and Ensemble Critic Set $\mathcal{Q} \leftarrow \emptyset$ \\
	Initialize: $\alpha_i \leftarrow 0.1$\\ 
	Initialize: $t \leftarrow 1.$ \\
	\textbf{repeat}
	\begin{algorithmic}[1]
		\State Observe state $\Phi$,  select the actions according to $\rho_{\textup{agent}} = \mu_{\theta} + \zeta $, $\rho_{\textup{adv}} = \nu_{\omega} + {\zeta}' $ where $\zeta$,${\zeta}' \sim \mathcal{N} \left ( 0,\sigma I\right )$.
		\State Execute the action $\rho_t = \alpha_i \cdot \rho_{\textup{adv}} + \left ( 1 - \alpha_i  \right ) \cdot \rho_{\textup{agent}} $ in the environment.
		\State Observe the reward $r$ in (\ref{eq:17}), obtain the next state ${\Phi}'$ according to (\ref{eq:12}) and check whether the UAV has reached its destination using the signal done $d$.
		\State Store $\left ( \Phi, \rho, r, {\Phi}',d \right )$ in the replay buffer $\mathcal{D}$.
		\State Reset the environment if the state ${\Phi}'$ is terminal.
		\If {its time to update agent and adversary model}
		\State $\omega_t, \omega^{\left ( 1 \right )}_t \leftarrow \omega_t, \theta^{\left ( 1 \right )}_t \leftarrow \theta_t$ 
		\For{$k=1,2,\dots ,K_t$}
		\State Sample a random minibatch of $N$ transitions $B =  \left \{ \left (\Phi, \bar{\rho}, r, {\Phi}', d  \right ) \right \}$ from $\mathcal{D}$.
		\State Compute the targets as per (\ref{eq:D6}).
		\State Compute the loss function to update the joint critic $\nabla_{\phi} L \left ( \phi \right )$ as per (\ref{eq:D7}).
		\State Compute the agent and adversary policy gradient according to (\ref{eq:D8}). 
		\State Compute the agent and adversary parameters as per (\ref{eq:D8}).
		\State Update the parameters of the agent and adversary network $\left \{ \theta ,\omega  \right \}$ according to (\ref{eq:D9}) and (\ref{eq:D10}).
		\State Obtain the value of $\left \{ \bar{\theta} ,\bar{\omega}  \right \}$ according to (\ref{eq:D11}).
		\State Update the parameters of the target network used to compute the target according to (\ref{eq:D9}).
		\begin{equation}\label{eq:D13}
			\begin{gathered}
				\phi_{\textup{targ}}  \leftarrow  \tau \phi_{\textup{targ}} + \left ( 1 - \tau \right )\phi \\
				\theta_{\textup{targ}}  \leftarrow  \tau \theta_{\textup{targ}} + \left ( 1 - \tau \right )\theta^{k+1}_t \\
				\omega_{\textup{targ}}  \leftarrow  \tau \omega_{\textup{targ}} + \left ( 1 - \tau \right )\omega^{k+1}_t
			\end{gathered}
		\end{equation}
		\EndFor
		
		\State Update $\left \{ \omega_{t+1}, \theta_{t+1}  \right \}$ as per (\ref{eq:D12}).
		\State $t\leftarrow t+1$
		\EndIf
		\State \textbf{until} Max Episodes Reached.
		\State Obtain $\Delta \mathbf{H}  = \mathbb{E} \left [ \Delta \mathbf{H}^{\textup{rand}}_{\alpha_i} - \max \left [ \mathbf{H}^{\textup{opt}}_{\alpha_i}  \right ]   \right ]  $.
		\If {$\Delta \mathbf{H} > \mathbf{H}^{\textup{th}}$}
		\begin{equation} \label{eq:D14}
			\begin{gathered}
				\alpha_i \leftarrow  \alpha_i + 0.1,\\
				\mathcal{P}' \leftarrow \mathcal{P} \cup \left\{ \theta, \omega \right\}_{\alpha_i}, \\
				\mathcal{Q}' \leftarrow \mathcal{Q} \cup \left\{ \phi \right\}_{\alpha_i}, \\
				\mathcal{P} \leftarrow \mathcal{P}', \\
				\mathcal{Q}' \leftarrow \mathcal{Q}.
			\end{gathered}
		\end{equation}  
		\Else 
		\space \textbf{End}
		\EndIf
	\end{algorithmic} 
	\textbf{Output:} Ensemble Policy set $\mathcal{P}$ and critic set $\mathcal{Q}$.
\end{algorithm}
\section{Algorithm for Generating Self-Induced Adversarial States}
\begin{algorithm}[H]
	\caption{Generate self-induced adversarial state $\Phi^{\textup{adv}}$ for obtaining the distributional shift from vanilla and robust policies}
	\label{alg:adv_state}  
	\textbf{Input} Current State $\Phi$, Step size $\alpha$, maximum perturbation $\epsilon$, mean observation $\breve{\mu}_{\Phi}$, standard deviation of observation $\breve{\sigma}_{\Phi}$.  \\
	\textbf{Output} Adversarial State $\Phi_{\textup{adv}}$
	\begin{algorithmic}[1]  
		\State Calculate normalized state $\Phi_{\textup{norm}} = \left ( \Phi - \breve{\mu}_{\Phi} \right ) \mathbin{/} \breve{\sigma}_{\Phi}.$
		\State Compute the perturbed input $\delta_0 = 2\epsilon \cdot  \mathcal{U} \left (-0.5,0.5 \right ) $.
		\For{ $k = 0, \cdots, N-1$}
		\State Compute the step size $\digamma = c \mathbin{/} k +c$.
		\State Obtained the perturbed state $\Phi_k = \Phi_{\textup{norm}} + \delta_0$.
		\State Obtain the gradient of the actor loss on perturbed state $g_k = \nabla_{\Phi_k} \mathcal{L}_{\pi} \left ( \Phi_k \right )$.
		\State Compute the perturbation step $s_k = \epsilon\cdot \textup{sign} \left (g_k \right ) $.
		\State Update the perturbation $\delta_{k+1} = \digamma_k s_k + \left ( 1 - \digamma_k \right ) \delta_k$.
		\EndFor
		\State Compute the final adversarial example in normalized form $\Phi^{\textup{norm}}_{\textup{adv}} = \Phi_{\textup{norm}} + \delta_N.$
		\State Renormalize to obtain the final adversarial state $\Phi_{\textup{adv}} = \breve{\sigma}_{\Phi} \cdot \Phi^{\textup{norm}}_{\textup{adv}} + \breve{\mu}_{\Phi}.$
	\end{algorithmic} 
\end{algorithm}
\section{Algorithm for Discounted Thompson Sampling}
The algorithm for DTS is provided as follows:
\begin{algorithm} [H]
	\caption{Discounted Thompson Sampling for Switching Robust Policies against Adversarial Attacks}
	\label{alg:dTS}  
	\textbf{Parameters} $\mathfrak{y} \in \left (0,1 \right ]$, $\mathfrak{a_0}, \mathfrak{b_0} \in \mathbb{R}_{\geq 0}$, $K = \left | \mathcal{K} \right | \geq 2 $  
	\begin{algorithmic}[1]  
		\State Initialize $\mathfrak{S}_k, \mathfrak{F}_k = 0$
		$\forall k \in \{1, \ldots, K\}$.
		\For{ $t = 1, \cdots, T$}
		\For{ $k = 1, \cdots, K$}
		$\Lambda_k \left (t \right ) \sim \mathfrak{B} \left ( \mathfrak{S}_k+ \mathfrak{a}_0, \mathfrak{F}_k+ \mathfrak{b}_0\right ). $
		\EndFor
		\State Play arm $I^{\varkappa}_t := \argmax_k \Lambda_k \left (t \right ) $ and observe the Bernoulli reward $\tilde{r}_t$ from the true probabilities $p_{\textup{true}}$.
		\State Perform a Bernoulli trial with success probability $\tilde{r}_t$ and observe output $\breve{r}_t$.
		\State Update $\mathfrak{S}_{I^{\varkappa}_t} \leftarrow \mathfrak{y} \mathfrak{S}_{I^{\varkappa}_t} + \breve{r}_t $ and $\mathfrak{F}_{I^{\varkappa}_t} \leftarrow \mathfrak{y} \mathfrak{F}_{I^{\varkappa}_t} + \left( 1-  \breve{r}_t \right )$.
		\State Update $\mathfrak{S}_{k} \leftarrow \mathfrak{y} \mathfrak{S}_k$ and $\mathfrak{F}_k \leftarrow \mathfrak{y} \mathfrak{F}_k ; \forall k \neq I^{\varkappa}_t$.
		\EndFor
	\end{algorithmic} 
\end{algorithm}
\section{Antifragility Proof} \label{AF:proof}
	By Theorem \ref{th:1},  the DTS switching strategy converges in probability to selecting the policy that minimizes the value distributional shift:
\begin{equation}
	\lim_{t \to \infty} \mathbb{P} \left( \kappa_t = \arg\min_k d_k^{(t)} \right) = 1.
\end{equation}
Let $k^* := \arg\min_k d_k^{(t)}$ denote the optimal policy under the current (unseen) perturbation.  Since DTS sampler utilizes the optimal policy $k^*$, and $r_k^{(t)} = d_{\min}^{(t)} / d_k^{(t)}$, we have:
\begin{equation}
	\lim_{t \to \infty} r_{\kappa_t}^{(t)} \to 1 \quad \Rightarrow \quad \lim_{t \to \infty} d_{\kappa_t}^{(t)} \to d_{\min}^{(t)}.
\end{equation}
By the Lipschitz assumption:
\begin{equation}
	\mathbb{E}[R(\pi_{\kappa_t}; \Phi_t^{\text{adv}})] \geq \mathbb{E}[R(\pi_0; \Phi_t)] - L \cdot d_{\kappa_t}^{(t)}.
\end{equation}
Differentiating both sides w.r.t.\ $t$, we get:
\begin{equation}
	\frac{d}{dt} \mathbb{E}[R(\pi_{\kappa_t}; \Phi_t^{\text{adv}})] \geq -L \cdot \frac{d}{dt} d_{\kappa_t}^{(t)}.
\end{equation}
As DTS converges to the optimal robust policy, $d_{\kappa_t}^{(t)}$ decreases over time:
\begin{equation}
	\frac{d}{dt} d_{\kappa_t}^{(t)} < 0 \quad \Rightarrow \quad \frac{d}{dt} \mathbb{E}[R(\pi_{\kappa_t}; \Phi_t^{\text{adv}})] > 0.
\end{equation}
\section{Benchmarking Algorithms} \label{bench_algo}
Benchmarking is performed by evaluating the standard policy gradient algorithm, such as the DDPG, along with its robust variants. An approach involves an agent who intends to execute an action $\mathbf{a}$, but an alternative adversarial action $\bar{\mathbf{a}}$ is implemented with probability $\alpha$. Another variant, conceptually similar to the approach described in Section \ref{adv_ml}, employs soft probabilistic robust policy iteration (PR-PI) as proposed in \cite{tessler2019action}, rather than solving the mixed Nash equilibrium objective in (\ref{eq:19}). The final benchmarking strategy involves training the agent to optimize the rewards without directly modifying the policy during training. Instead, an adversarial buffer is utilized, allowing the agent to learn the optimal value function in the presence of adversarial states. A brief explanation of these algorithms is provided in the subsequent subsections.
\subsection{Deep Deterministic Policy Gradient (DDPG)}
The schematic shown in Figure \ref{figure2_schematic}, the action robust RL is replaced by DDPG considering $\alpha = 0$. Hence, the critic target $y_{\textup{targ}}$ in (\ref{eq:D6}) is modified as follows,
\begin{equation} \label{eq:H1}
	y_{\textup{targ}} = r_t + \gamma \left (1-d \right ) Q_{\phi_{\textup{targ}}} \left ( \Phi', \mu_{\theta}  \right ).
\end{equation}
Similarly the policy loss with $\alpha = 0$ shown in (\ref{eq:D8}) to compute the policy parameter $\theta$, will be as,
\begin{equation} \label{eq:H2}
	\nabla_{\theta} \widehat{J \left ( \theta\right )} = \frac{1}{N}  \sum_{\Phi \in \mathcal{D}} \nabla_{\theta} \mu_\theta \left ( \Phi \right )  \nabla_a Q_{\phi} \left ( \Phi, a \right ).
\end{equation}
Hence, the min-max objective in (\ref{eq:19}) is replaced with a pure max objective which can be solved using policy gradient methods as outlined in \cite{duan2016benchmarking}. 

\subsection{Probabilistic Robust Markov Decision Process}
According to \cite{tessler2019action}, PR-MDP is framed as a zero-sum game between an agent and an adversary. In this framework, an optimal probabilistic robust policy is defined with a probability $\alpha$, where the adversary assumes control and executes the most detrimental actions to account for potential system control limitations and undesirable outcomes. PR-MDP addresses stochastic perturbations within the policy space.  When we consider $\alpha \in \left [ 0, 1 \right ]$ and the 5-tuple MDP outlined in Section \label{adv_ml}, the probabilistic joint policy $\pi^{\textup{mix}}_{P,\alpha}$ is considered as 
\begin{equation} \label{eq:51}
	\pi^{\textup{mix}}_{P,\alpha} \left ( \left.\begin{matrix}
		\mathbf{a} 
	\end{matrix}\right| \Phi  \right )  \equiv \left ( 1 - \alpha \right ) \pi \left ( \left.\begin{matrix}
		\mathbf{a} 
	\end{matrix}\right| \Phi \right ) + \alpha \bar{\pi} \left ( \left.\begin{matrix}
		\mathbf{a} 
	\end{matrix}\right| \Phi \right ). 
\end{equation}
The $\pi^{\textup{mix}}_{P, \alpha}$ to learn the optimal value function for PR-MDP is obtained using PR-PI as shown in algorithm 1 in \cite{tessler2019action}.

\subsection{Noisy Robust Markov Decision Process}
According to \cite{tessler2019action}, NR-MDP also models a zero-sum game between an agent and an adversary, but incorporates perturbations in the action space rather than in the policy space as in PR-MDP. Given $\pi$ and $\bar{\pi}$ as the policies of the agent and the adversary, respectively, their joint policy $\pi^{\textup{mix}}_{P,\alpha}$ as, 
\begin{equation} \label{eq:52}
	\pi^{\textup{mix}}_{N, \alpha} \left ( \left.\begin{matrix} \mathbf{a}  \end{matrix}\right| \Phi \right ) 
	= \mathbb{E}_{\mathbf{b} \sim \pi \left ( .| \Phi \right ),  \bar{\mathbf{b}} \sim \bar{\pi} \left ( .| \Phi \right )} \left [ \vec{\mathbf{1}}_{\mathbf{a} = \left ( 1 - \alpha \right ) \mathbf{b} + \alpha \bar{\mathbf{b}} } \right ].
\end{equation}
The NR-MDP is analogous to the Markov decision process (MDP) used to develop the ensembles, except for the addition of SGLD noise to the loss function, as described in equation (\ref{eq:D4}) to obtain the mixed Nash equilibrium solution for the objective shown in (\ref{eq:19}). The solution to the min-max objective considered here is according to the soft PR-PI method as described in Algorithm 2 in \cite{tessler2019action}.
\subsection{Learning with Adversarial Attacks}
Adversarial attacks, as proposed in \cite{pattanaik2018robust}, have been used to train DRL algorithms to address environmental uncertainties. These attacks employ the fast gradient sign method (FGSM) to introduce perturbations in the observation space. In the case of DDPG, attacks are applied considering the gradient of the policy loss function. Adversarial training is conducted within a robust control framework, where the agent is trained to handle state-space perturbations that could potentially cause the greatest impact on the reward space. The primary distinction between vanilla DDPG and adversarial DDPG lies in the use of adversarial states by the latter in the training buffer, represented as $\mathcal{D}^{\textup{adv}} =\left\{ \Phi^{\textup{adv}}, a^{\textup{adv}}, \Phi', r^{\textup{adv}} \right\}$, where the adversarial state is given by $\Phi^{\textup{adv}} = \Phi + \epsilon \cdot \textup{sign} \left [ \nabla_{\Phi} J\left ( \theta \right ) \right ]$. Here, $J$ denotes the loss of policy obtained from DDPG. In the simulation, the adversarial DDPG model was trained with $\epsilon = 1.0$.

\end{document}